\documentclass[10pt,journal,compsoc]{IEEEtran}
\usepackage{times}
\usepackage{epsfig}
\usepackage{graphicx}
\usepackage{tabularx}
\usepackage{adjustbox}
\usepackage{amsmath}
\usepackage{amssymb}
\usepackage{multirow}
\usepackage{xcolor}
\usepackage{booktabs}
\usepackage{soul}
\usepackage{subcaption}
\usepackage{xspace}
\usepackage{mathrsfs}
\usepackage{soul}

\usepackage{array}
\usepackage{tabu}
\usepackage[hang,flushmargin,norule]{footmisc}
\renewcommand{\footnoterule}{%
  \kern -3pt 
  \hrule width 0.4\textwidth height 0.4pt 
  \kern 2pt 
}
\usepackage{color}
\usepackage{xcolor}
\usepackage{hyperref}
\usepackage{flushend}

\ifCLASSOPTIONcompsoc
  \usepackage[nocompress]{cite}
\else
  \usepackage{cite}
\fi
\ifCLASSINFOpdf
\else
\fi
\begin{document}
%
\title{Global and Local Semantic Completion Learning for Vision-Language Pre-training}
%
%
%
%

\author{Rong-Cheng Tu$^*$, Yatai Ji$^*$, Jie Jiang$^*$, Weijie Kong, Chengfei Cai, Wenzhe Zhao, Hongfa Wang, Yujiu Yang$^\dagger$, and Wei Liu$^\dagger$, \IEEEmembership{Fellow,~IEEE}
\IEEEcompsocitemizethanks{
\IEEEcompsocthanksitem Yatai Ji, Yujiu Yang are with Shenzhen International Graduate School, Tsinghua University, Shenzhen 518055, P.R.~China.(e-mail: jyt21@mails.tsinghua.edu.cn; yang.yujiu@sz.tsinghua.edu.cn)
\IEEEcompsocthanksitem Jie Jiang, Weijie Kong, Chengfei Cai, Wenzhe Zhao, Hongfa Wang, and Wei Liu are with the Tencent Data Platform, Shenzhen, Guangdong, China. (e-mail: \{zeus, jacobkong, fletchercai, carsonzhao, hongfawang\}@tencent.com;  wl2223@columbia.edu)
\IEEEcompsocthanksitem Rong-Cheng Tu, Yatai Ji and Jie Jiang are co-first authors. \protect  Work was done when Rong-Cheng Tu was
			interning at Tencent,  and his e-mail  is turongcheng@gmail.com.
\IEEEcompsocthanksitem Corresponding author: Yujiu Yang and Wei Liu. }
}

%
%

\markboth{IEEE Transactions on Pattern Analysis and Machine Intelligence}%
{Shell \MakeLowercase{\textit{et al.}}: Bare Demo of IEEEtran.cls for Computer Society Journals}
%



\IEEEtitleabstractindextext{%
\begin{abstract}
Cross-modal alignment plays a crucial role in vision-language pre-training (VLP) models, enabling them to capture meaningful associations across different modalities. 
For this purpose, inspired by the success of masked language modeling (MLM) tasks in the NLP pre-training area, numerous masked modeling tasks have been proposed for VLP to further promote cross-modal interactions. 
The core idea of previous masked modeling tasks is to focus on reconstructing the masked tokens based on visible context for learning local-local alignment, i.e., associations between image patches and text tokens.
However, most of them pay little attention to the global semantic features generated for the masked data, resulting in a limited cross-modal alignment ability of global representations to local features of the other modality. 
Therefore, in this paper, we propose a novel \textbf{G}lobal and \textbf{L}ocal \textbf{S}emantic \textbf{C}ompletion \textbf{L}earning (GLSCL) task to facilitate global-local alignment and local-local alignment simultaneously. 
Specifically, the GLSCL task complements the missing semantics of masked data and recovers global and local features by cross-modal interactions.
Our GLSCL consists of masked global semantic completion (MGSC) and masked local token completion (MLTC).
MGSC promotes learning more representative global features, which have a great impact on the performance of downstream tasks, while MLTC reconstructs modal-fusion local tokens, further enhancing accurate comprehension of multimodal data. 
To evaluate the proposed approaches on cross-modal alignment, we develop a validation benchmark called ALIGN-BENCH. 
Moreover, we present a flexible vision encoder, enabling our model to simultaneously perform image-text and video-text multimodal tasks. 
Experimental results show that our proposed method obtains state-of-the-art performance on various vision-language benchmarks, such as visual question answering, image-text retrieval, and video-text retrieval.
\end{abstract}

}

\maketitle

\IEEEdisplaynontitleabstractindextext

%
\IEEEpeerreviewmaketitle


\IEEEraisesectionheading{\section{Introduction}\label{sec:introduction}}

\IEEEPARstart{O}{ur} real-world contains a wide variety of information, such as texts, images, sounds, etc. 
To develop a powerful general artificial intelligence system, it is necessary to capture the semantic associations from different modality sources. 
Towards this goal, multimodal representation learning emerges as a critical technique for bridging the heterogeneity gap between different modalities~\cite{DBLP:journals/corr/abs-2109-04290/dualsoft, DBLP:journals/corr/abs-2206-08916/unifiedio}. 
In this field, vision-language pre-training models \cite{DBLP:conf/emnlp/TanB19/lxmert,DBLP:conf/nips/LuBPL19/vilbert,DBLP:conf/acl/LiGNXLL0020/unimo,DBLP:conf/cvpr/GeGLLSQL22/mcq,DBLP:journals/corr/abs-2210-05335/map} have shown an impressive semantic alignment ability,  which brings substantial advances on various downstream tasks, for instance, visual question answering, image-text retrieval, etc.

\begin{figure}[tp]
\centering
  \includegraphics[width=0.5\textwidth]{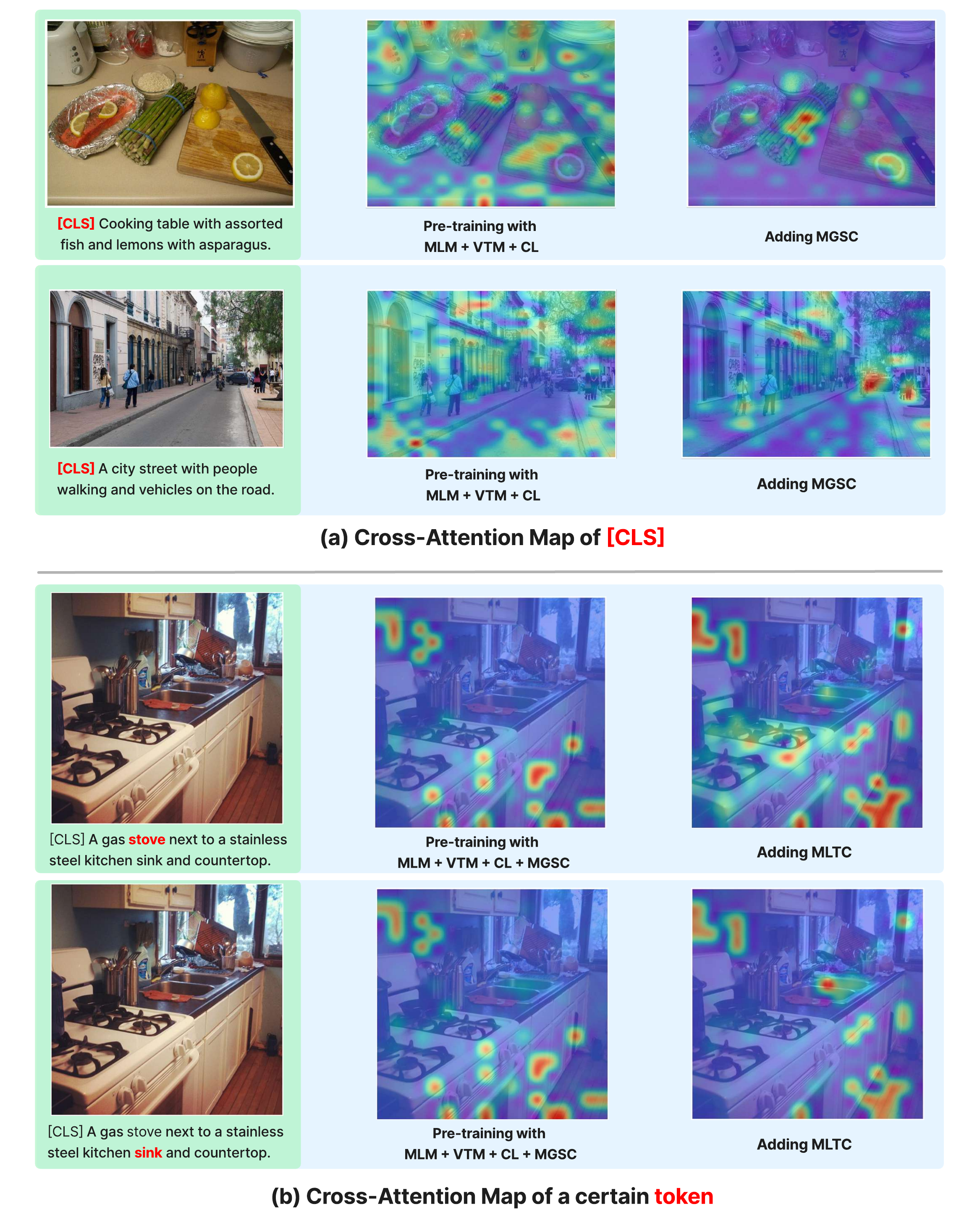}
  \caption{(a) The cross-modal attention map visualization of the text global representation ([CLS]) on the input image for our model pre-trained with or without MGSC. (b) The cross-modal attention map visualization of text local tokens for our model pre-trained with or without MLTC. The selected token is in \textcolor{red}{red}.}
  \label{fig:intro_visulize}
\end{figure}

Recently, numerous self-supervised vision-language pre-training models~\cite{DBLP:conf/emnlp/TanB19/lxmert, DBLP:conf/eccv/Li0LZHZWH0WCG20/oscar, fu2021violet, DBLP:conf/cvpr/GeGLLSQL22/mcq, bain2021frozen,DBLP:journals/corr/abs-2208-09374/vlmae, DBLP:journals/corr/abs-2206-01670/egocentric} have been proposed. 
These methods model the interactions between vision and language features mainly by using various masked modeling tasks, such as masked language modeling (MLM) and masked vision modeling (MVM). 
The basic idea of MLM and MVM is self-reconstructing the masked tokens via leveraging informative visible tokens to realize local-local alignment. 
Specifically, MLM adopted by BERT~\cite{kenton2019bert} is to predict the original vocabulary IDs of the masked words. 
Inspired by the success of MLM in pre-training, there is a flourishing trend to extend it to visual pre-training tasks. 
Generally, by masking some visual patches, MVM tasks predict their original pixels~\cite{DBLP:journals/corr/abs-2208-09374/vlmae, DBLP:conf/mm/Gao0LGWLY22/calic}, corresponding discrete tokens~\cite{DBLP:journals/corr/abs-2206-01127/vlbeit, DBLP:journals/corr/abs-2208-10442/beit3, fu2021violet} generated by the VQ-VAE variants, or Histograms of Oriented Gradients (HOG) features~\cite{DBLP:journals/corr/abs-2209-01540/violet2}, etc.

These existing masked modeling tasks only focus on reconstructing the local masked tokens and pay little attention to recovering the missing global semantic information caused by data corruption. 
The token-level reconstruction may lead to inadequate learning of global representations for cross-modal information. 
As illustrated in Fig.~\ref{fig:intro_visulize}(a), in the situation of token-level reconstructions (the middle column), the global representation exhibits disordered attention towards the other modality. 
It implies that the global-local cross-modal alignment ability of the pre-training model is limited, leading to a degraded global representation. 
However, the global semantic features have a great impact on the performance of the pre-training model as they are usually used to deal with downstream tasks. 
Therefore, it is crucial to ensure the global semantic features to learn more accurate global-local alignment. 

Therefore, in this paper, we propose a novel pre-training task called \textbf{G}lobal and \textbf{L}ocal \textbf{S}emantic \textbf{C}ompletion \textbf{L}earning (GLSCL), which recovers the global features and masked token features simultaneously. 
Our proposed GLSCL is based on the consideration that the paired vision and text data are two views of the same semantic information, allowing the missing semantics of masked data to be completed by capturing information from the other modality. 
Specifically, the GLSCL consists of two sub-tasks: masked global semantic completion (MGSC) and masked local token completion (MLTC).

The MGSC sub-task exploits information from complete text (vision) data to recover the global semantic representations of masked vision (text) data and is adapted to learn representative global features with accurate global-local alignment. 
For example, as illustrated in Fig.~\ref{fig:intro_visulize}(a), compared to the model pre-trained without MGSC, the attention maps with MGSC pre-training are more discriminative and reasonable. 
On the other hand, the MLTC sub-task focuses on learning more accurate local-local alignment to reconstruct the local masked tokens. 
As shown in Fig.~\ref{fig:intro_visulize}(b), after pre-training with MLTC, the model can attend to the relevant image regions corresponding to the words. 
Although the existing MVM and MLM tasks also involve reconstructing locally masked tokens, our proposed MLTC differs significantly from them. 
The MVM and MLM tasks solely rely on their respective original single modal signals, such as pixels or token IDs, as guiding information for reconstructing the masked tokens. 
In contrast, our proposed MLTC leverages the semantic features of completed data generated by the fusion encoder as supervised information. 
These features are enhanced through multimodal interactions, containing abundant intra- and inter-modal semantic information. 
Consequently, by reconstructing the masked tokens using our MLTC task, our model will be optimized comprehensively and sufficiently to generate representative features. 
Additionally, aiming at measuring the cross-modal global-local and local-local alignment of our model, we further manually annotate a validation benchmark ALIGN-BENCH to conduct quantitative experiments.


For the architecture of the vision-language pre-training model, we adopt a general framework that consists of two uni-modal encoders and a fusion encoder. 
Moreover, we present a flexible vision encoder to enable our model to perform image-text and video-text multimodal tasks simultaneously. Specifically, for video inputs, the vision encoder only adds a few additional learning parameters and the [CLS] feature of each frame is treated as a bridge associating spatial modeling within the frame and temporal modeling among frames. 
Inspired by curriculum learning~\cite{bain2021frozen}, we train the model with image-text and video-text datasets successively to transfer visual knowledge from images to videos.

In a nutshell, our contributions are four-fold. 
(1) To enhance the global-local alignment of global representations, we propose a new pre-training task called masked global semantic completion (MGSC), which recovers missing semantic information from unmasked data, promoting learning more representative global features. 
(2) Furthermore, we propose masked local token completion (MLTC) to enhance the local-local alignment of vision and text tokens. 
(3) We manually annotate a validation benchmark ALIGN-BENCH to quantitatively demonstrate the effectiveness of the proposed GLSCL on cross-modal alignment. 
(4) We design an adaptive vision encoder, which can transfer multimodal pre-training knowledge between images and videos readily. 
(5) We conduct multiple vision-language downstream tasks to demonstrate the generalization of global and local semantic completion learning (GLSCL), including visual question answering, visual reasoning, image-text retrieval, and video-text retrieval. 
Our model achieves state-of-the-art performance based on a similar pre-training data scale. 
Compared to some existing similar models, our model also gains the best results on ALIGN-BENCH. 
Our code is available at https://github.com/IIGROUP/SCL.

\section{Related Work}
\subsection{Vision-Language Pre-training}
Vision-language pre-training is an important way to learn effective universal multimodal representations. 
Existing vision-language pre-training frameworks can be divided into two categories: dual-tower and cross-fusion architecture.  

The dual-tower architecture-based methods \cite{DBLP:conf/icml/RadfordKHRGASAM21/clip,jia2021scaling,gabeur2020multi,xu2021videoclip,bain2021frozen,akbari2021vatt} employ two individual encoders to separately extract the features for the visual data (images or videos) and textual data, and then map these features into a common semantic space. Among them, CLIP \cite{DBLP:conf/icml/RadfordKHRGASAM21/clip} exploits contrastive learning with a huge quantity of noisy image-text pairs directly collected from the Internet, achieving remarkable results and strong generalization on plenty of vision-language tasks~\cite{DBLP:journals/corr/abs-2204-06125/dalle2, DBLP:conf/cvpr/WangLLTGGL22/cris, DBLP:journals/corr/abs-2212-04873/prototype}. 
To further improve the CLIP \cite{DBLP:conf/icml/RadfordKHRGASAM21/clip}, Alex et al.~\cite{DBLP:conf/cvpr/AndonianCH22} proposes to use progressive self-distillation and soft image-text alignments to efficiently learn robust representations from noisy data.
COTS \cite{DBLP:conf/cvpr/LuFHG0W22} leverages token-level interaction, instance-level interaction and task-level interaction to model fine-grained interactions between the vision and text modalities.
The aforementioned methods are proposed for dealing the image-text data, for video-text data, there are also a lot of works.  VATT \cite{akbari2021vatt} employs multimodal contrastive learning to align the video frames, audios and texts, and achieves impressive performance on the downstream tasks. FROZEN \cite{bain2021frozen} proposes a curriculum learning schedule to train the vision-language model on both image-text and video-text datasets by treating an image as a single-frame video. MILES \cite{DBLP:conf/eccv/GeGLWWSQL22} exploits masked visual modeling in video-text pre-training with the “dual-encoder” architecture, which adopts a snapshot video encoder to produce reconstruction targets with injected language semantics for the masked video patch prediction.
Although these dual-tower architecture based methods perform well on cross-modal retrieval tasks with high efficiency, their performances on the more complex multimodal downstream tasks are not inspirational due to the insufficient interaction between vision and text features.

\begin{figure*}[tp]
	\centering
	\includegraphics[width=0.9\textwidth]{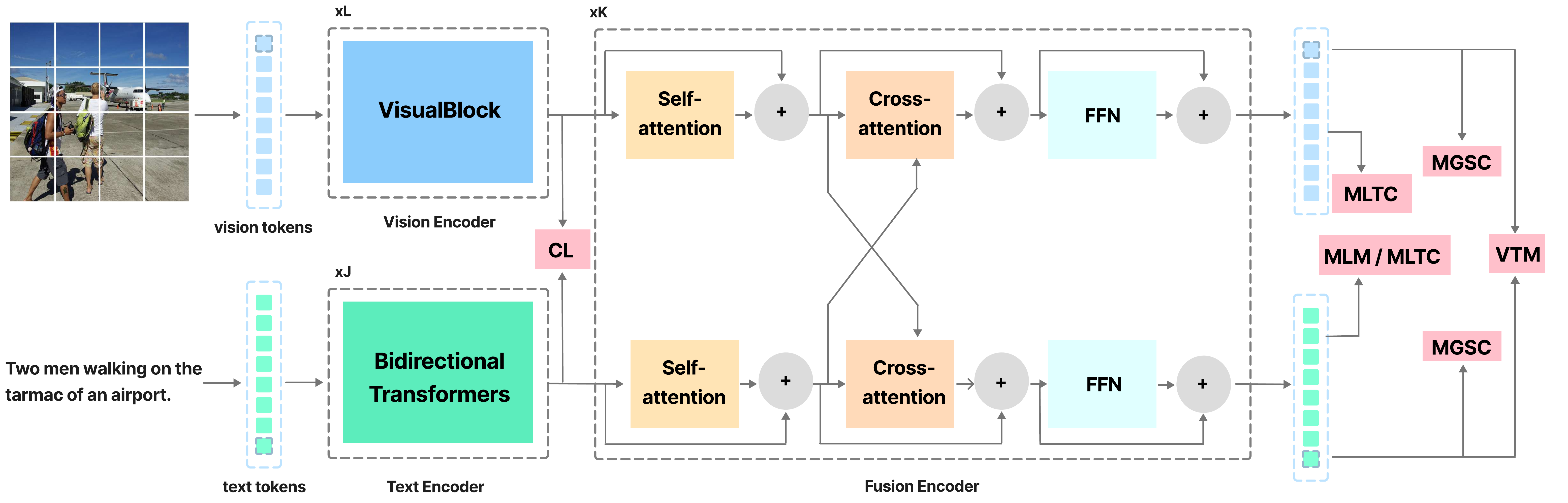}
	\centering
	\caption{The whole architecture of our model.}
	\label{fig:architecture}
 \vspace{-0.2cm}
\end{figure*} 
To overcome this limitation, the cross-fusion architecture based methods~\cite{DBLP:conf/emnlp/TanB19/lxmert, DBLP:conf/acl/LiGNXLL0020/unimo, chen2020uniter, DBLP:conf/eccv/Li0LZHZWH0WCG20/oscar, Dou_2022_CVPR/meter} have been proposed, which employ a cross-modal fusion encoder to enhance the interactions between vision and text features. For the image-text scenarios,  ALBEF \cite{li2021albef} not only aligns the image and text features with contrastive learning but also feeds them into a cross-modal attention-based encoder to obtain the fused features. 
TCL \cite{DBLP:conf/cvpr/YangDTXCCZCH22/tcl} also employs the aligning-before-fusing approach proposed by ALBEF, but takes it a step further by simultaneously utilizing cross-modal and intra-modal self-supervision during pre-training. This method enhances multi-modal interactions, resulting in the most recent state-of-the-art performance in multi-modal tasks.
ViLT \cite{DBLP:conf/icml/KimSK21/vilt} proposes a minimal VLP architecture that adopts simple unimodal encoders to deal with texts and images, and focuses more on the modality interactions inside the fusion module. 
Moreover, VLMo~\cite{DBLP:journals/corr/abs-2111-02358/vlmo} and Beitv3~\cite{DBLP:journals/corr/abs-2208-10442/beitv3} adopt modality-specific FFNs and a shared self-attention layer in each block to conduct single-modality learning and cross-modal interactions flexibly. 
Similarly, for the video-text data, Clover~\cite{DBLP:journals/corr/abs-2207-07885/clover} improves video-text feature alignment and fusion via a tri-modal alignment pre-training task. 
ALPRO \cite{DBLP:conf/cvpr/Li0LNH22/alpro} proposes a novel approach to aligning instance-level unimodal features by introducing video-text contrastive learning, and by utilizing prompt entity modeling to achieve fine-grained region-entity alignment.
In this work, to achieve superior performance in downstream tasks, we also adopt the "cross-fusion" architecture to improve the interaction between the visual and textual modalities.
\subsection{Masked Modeling Tasks}
Recently, various masked modeling tasks have been proposed, whose strategy is self-reconstructing the masked data. Masked Language modeling (MLM) inspired by the Cloze task~\cite{taylor1953cloze} is the most classical one. It randomly masks some tokens of the input and then predicts the original vocabulary IDs of the masked words based on their context. 
By pre-training with the MLM, BERT~\cite{kenton2019bert} achieved state-of-the-art results on eleven natural language processing (NLP) tasks. 
Afterward, for promoting cross-modal interaction, MLM is widely adopted in vision-language pre-training models~\cite{DBLP:conf/emnlp/TanB19/lxmert, li2021albef, DBLP:conf/emnlp/WangJSYS21/mirtt, Dou_2022_CVPR/meter}. 
Inspired by the success of MLM, some works extend it into the visual domain and propose masked vision modeling (MVM). After masking some visual patches, some MVM works mainly to predict the corresponding discrete tokens~\cite{DBLP:journals/corr/abs-2206-01127/vlbeit, DBLP:journals/corr/abs-2208-10442/beit3, fu2021violet} generated by VQ-VAE or its variants; other works directly regress the original pixels~\cite{DBLP:journals/corr/abs-2208-09374/vlmae, DBLP:journals/corr/abs-2208-02131/maevl} or Histograms of Oriented Gradients (HOG) features~\cite{DBLP:journals/corr/abs-2209-01540/violet2} of masked visual tokens. 
For example, VIOLET \cite{fu2021violet} proposes a masked visual-token modeling task, which first maps the original video frame patches into discrete visual tokens and then recovers the corresponding visual tokens of masked patches to train a joint encoder for the vision-language fusion. 
VLMAE~\cite{DBLP:journals/corr/abs-2208-09374/vlmae} proposes the Regional Masked Image Modeling (RMIM) task to facilitate the fusion of multimodal features. The RMIM masks some patches of an input image and then reconstructs the original pixels depending on the visible patches and the corresponding text. 
MaskFeat \cite{DBLP:conf/cvpr/00050XWYF22} randomly masks a portion of a video sequence and then regresses the HOG features of the masked regions.

However, these tasks focus on reconstructing local masked tokens, ignoring the recovery of global semantic information of the masked data after cross-modal interactions. 
Hence, we propose a novel masked global semantic completion (MGSC) task. 
Moreover, our method also recovers semantic representations of masked tokens for further cross-modal alignment of local features, i.e., masked local token completion (MLTC).

\section{Approaches}

In this section, we first introduce our pre-training objectives in Sec.~\ref{pre-training task} and then describe the model architecture in Sec.~\ref{model architecture}. 
The whole architecture is displayed in Fig.~\ref{fig:architecture}.


\subsection{Pre-training Tasks}
\label{pre-training task}
\noindent\textbf{Contrastive Learning (CL).} 
The input images and texts are projected into vision and language embedding spaces with two uni-modal encoders, respectively. 
We utilize contrastive learning to adjust the positions of semantic features, enforcing the paired image-text features close and negative samples far apart. 
Specifically, we conduct CL on the global representations from vision and text encoders. 
Given a batch of image-text pairs, for an image (text), the paired text (image) is treated as the positive sample, and other texts (images) are negative samples. 
We use the InfoNCE loss as follows:
\begin{equation}
\begin{aligned}
\operatorname{\textbf{NCE}}_{V2T}=-\frac{1}{N}\sum_{i=1}^{N}\operatorname{log}\frac{\operatorname{exp}(s(V_i,T_i)/\tau)}{\sum_{n=1}^N\operatorname{exp}(s(V_i,T_n)/\tau)}\, , \\
\operatorname{\textbf{NCE}}_{T2V}=-\frac{1}{N}\sum_{i=1}^{N}\operatorname{log}\frac{\operatorname{exp}(s(T_i,V_i)/\tau)}{\sum_{n=1}^N\operatorname{exp}(s(T_i,V_n)/\tau)}\, ,
\end{aligned}
\label{eq:cl nce loss}
\end{equation}
where $N$ is the batchsize and $\tau$ serves as a learnable temperature parameter. 
The similarity function is formatted as cosine similarity, $s(V,T)=\frac{\phi_v(V)^T\phi_t(T)}{||\phi_v(V)||\cdot ||\phi_t(T)||}$, where $\phi$ is a linear projection head. 
The vision-text contrastive loss is defined as:
\begin{equation}
\mathcal{L}_{CL} = \operatorname{\textbf{NCE}}_{V2T} +\operatorname{\textbf{NCE}}_{T2V}.
\end{equation}

For the video-text data, we use the mean pooling of $M$ frame [CLS] features to denote the global representation of a video and then also use Eq. (\ref{eq:cl nce loss}) for contrastive learning. 
Afterward, token-wise fusion is employed for features of different modalities in the unified semantic space. 

\noindent\textbf{Vision Text Matching (VTM).} 
VTM aims to determine the correspondence of an image-text pair. 
The model conducts a binary classification on the concatenation of vision and text global representations generated by the fusion encoder, which contributes to the overall alignment of different modalities. 
The loss is defined as:
\begin{equation}
\mathcal{L}_{VTM}=\operatorname{\textbf{CE}}(\phi(\operatorname{concat}[V, T]),y),
\end{equation}
where $V$, $T$ are [CLS] features and $y$ is the ground truth. 
$\operatorname{\textbf{CE}}$ is Cross-Entropy loss and $\phi$ refers to a binary classifier. 
The image-text (video-text) pairs serve as positive samples, and we randomly replace the image (video) in a data pair with another image (video) to build a negative sample.

\noindent\textbf{Masked Language Modeling (MLM).} 
MLM was first used as a pretext task in natural language processing~\cite{taylor1953cloze, kenton2019bert} and was later introduced to multi-modal pre-training. 
Following the text tokens masked out, the model attempts to predict the original words based on visual information and textual context. 
We randomly mask 15$\%$ text tokens and replace them with the [MASK] token, random words, or left unchanged, with the probability of 80$\%$, 10$\%$, and 10$\%$, respectively. 
Then, the model conducts a classification on the vocabulary list to predict masked words. 
The classification loss is as follows:
\begin{equation}
\mathcal{L}_{MLM} = \operatorname{\textbf{CE}}(\phi(T_{mask}),y),    
\end{equation}
where $T_{mask}$ is the output masked token feature, $\phi$ serves as a classifier, and $y$ is the original token ID. 
The token-level reconstruction task plays an important role in the way that the model learns to associate linguistic words and visual entities, realizing local-local semantic alignment. 

\begin{figure*}[tp]
\centering
  \includegraphics[width=\textwidth]{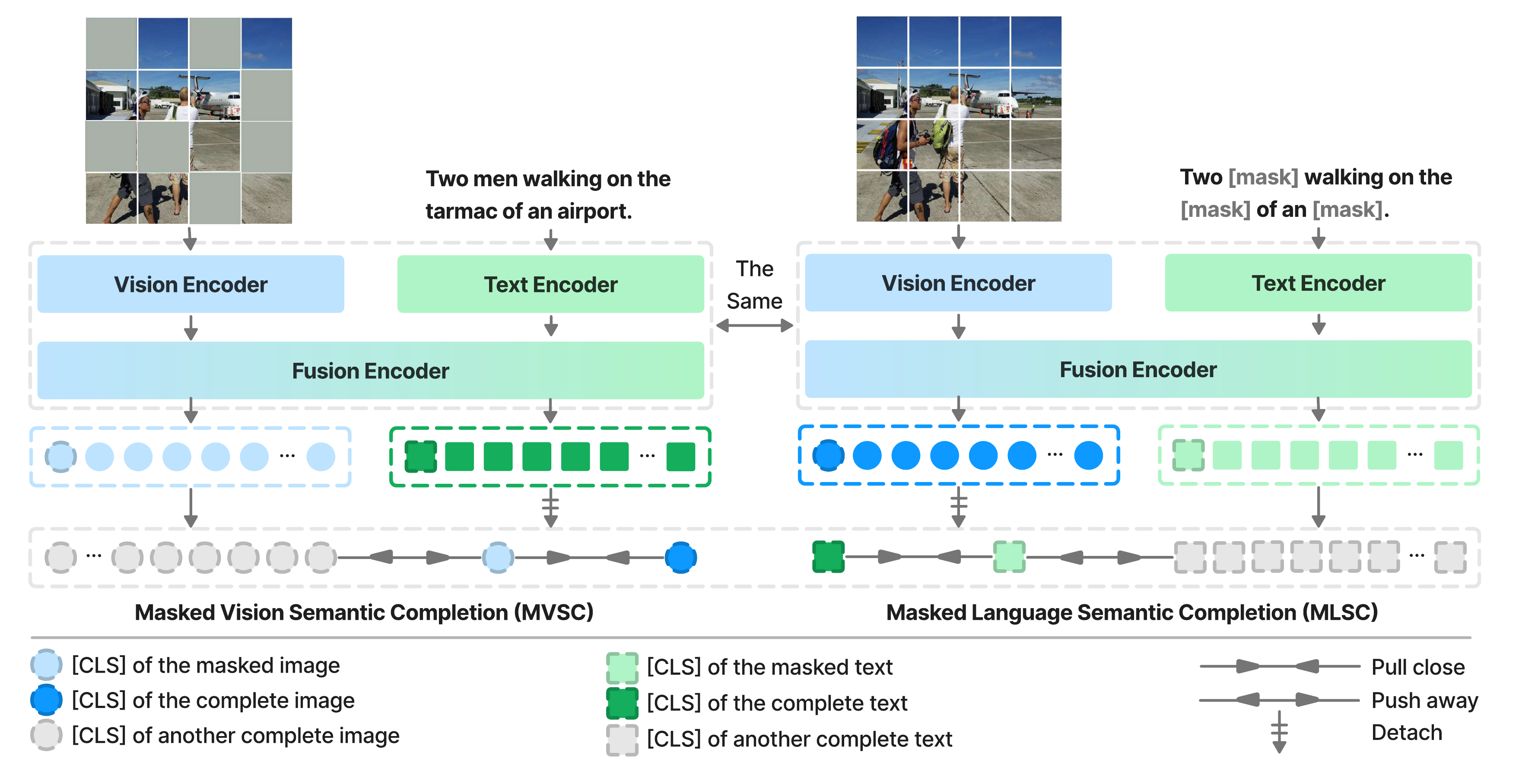}
  \caption{The overview of our proposed Masked Global Semantic Completion (MGSC). The two versions of an image-text pair are forward propagated, respectively, to perform masked vision/language semantic completion.}
  \label{fig:SCL}
  \vspace{-0.2cm}
\end{figure*}

\noindent\textbf{Semantic Completion Learning (SCL).} 
It is significant for the model to learn multi-modal information fusion, that is, to extract knowledge from the other modality. 
To this end, we expect that the model can recover the global and local semantics of masked images or texts after cross-modal interaction, enabling accurate alignment between global and local features with tokens of the other modality. 
In this way, the model can leverage knowledge from one modality to enhance the understanding and representation of the other modality.

\begin{figure}[tp]
\centering
  \includegraphics[width=0.5\textwidth]{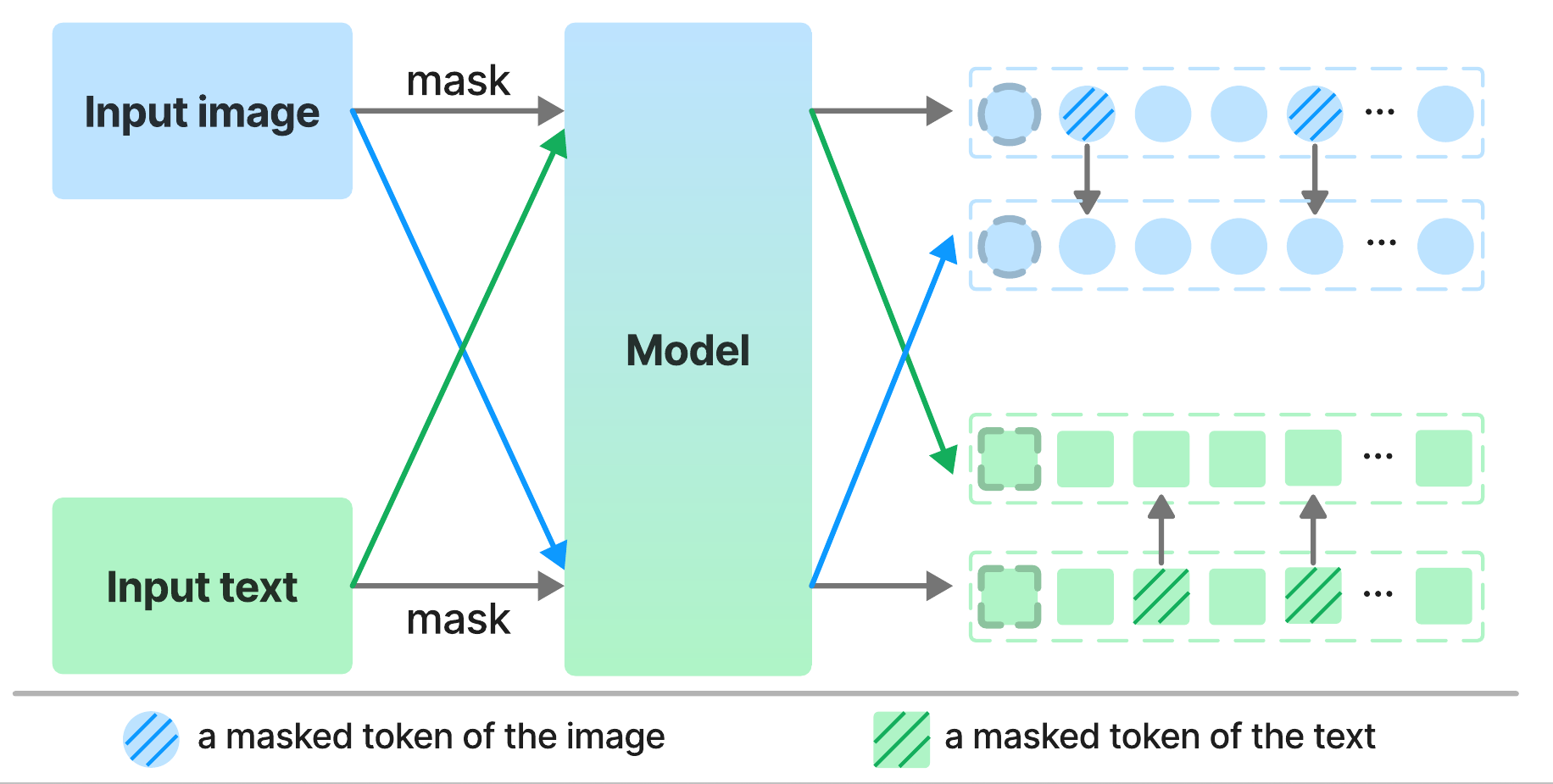}
  \caption{The overview of our proposed Masked Local Token Completion (MLTC).}
  \label{fig:token reconstruction}
  \vspace{-0.3cm}
\end{figure}

\textbf{M}asked \textbf{g}lobal \textbf{s}emantic \textbf{c}ompletion (MGSC) is shown in Fig. \ref{fig:SCL}. For each data pair, we first randomly mask the image and text separately to get $\{I_{mask}, T\}$ and $\{I, T_{mask}\}$, so that the masked one manages to learn semantic information from the other complete modality.  
Then the two couples of data are sent to the model respectively. 
The recovered features of masked data are obtained by leveraging information from the other modality to complete its missing semantic information: 
\begin{equation}
\begin{aligned}
I_{Re}, T_{Co}&=\operatorname{Model}(I_{mask}, T), \\
I_{Co}, T_{Re}&=\operatorname{Model}(I, T_{mask}), \\
\end{aligned}
\label{msm_forward}
\end{equation}
where $I_{Re}, T_{Re}$ are \textbf{re}covered global features of masked data, and $I_{Co}, T_{Co}$ refer to global features of \textbf{co}mplete data. 
Then, we conduct masked vision semantic completion (MVSC) and masked language semantic completion (MLSC) simultaneously. 
Specifically, we bridge the gap between recovered global features and the complete ones in the form of contrastive learning. 
The InfoNCE loss is adopted to maximize the mutual information (MI) between two versions of the input data pair, $\{I_{mask}, T\}$ and $\{I, T_{mask}\}$: 
\begin{equation}
\begin{aligned}
\operatorname{\textbf{NCE}}_V=-\frac{1}{N}\sum_{i=1}^{N}\operatorname{log}\frac{\operatorname{exp}(s(I_{Re}^i,I_{Co}^i)/\tau)}{\sum_{n=1}^N\operatorname{exp}(s(I_{Re}^i,I_{Co}^n)/\tau)}\, , \\
\operatorname{\textbf{NCE}}_L=-\frac{1}{N}\sum_{i=1}^{N}\operatorname{log}\frac{\operatorname{exp}(s(T_{Re}^i,T_{Co}^i)/\tau)}{\sum_{n=1}^N\operatorname{exp}(s(T_{Re}^i,T_{Co}^n)/\tau)}\, ,
\end{aligned}
\label{eq:nceloss}
\end{equation}
where $s$ denotes cosine similarity and $\tau$ serves as the temperature hyper-parameter. 
The negative samples are global features of other complete images or texts in a batch. 
Note that $I_{Co}$ and $T_{Co}$ are detached for gradient backward, which makes the model more focused on the learning of recovering global features. 
Finally, the global semantic completion learning loss is defined as: 
\begin{equation}
\mathcal{L}_{MGSC}=\operatorname{\textbf{NCE}}_V+\operatorname{\textbf{NCE}}_L.
\label{eq:scl}
\end{equation}

By minimizing the Eq.(\ref{eq:scl}), it will make the global feature $I_{Re}$ of the masked image similar to $I_{Co}$ of the complete image ($T_{Re}$ similar to $T_{Co}$). 
To recover the semantic information of masked data, the global representations will learn supplementary knowledge from corresponding tokens of the other modality, i.e., accurate global-local alignment. 

As shown in Fig. \ref{fig:token reconstruction}, the \textbf{m}asked \textbf{l}ocal \textbf{t}oken \textbf{c}ompletion (MLTC) is similar to the aforementioned process. 
We recover token features of masked data with contrastive loss to complete local semantic information, facilitating local-local cross-modal alignment. 

Specifically, the masked positions of images or texts are replaced with learnable [MASK\_V] or [MASK\_L] tokens. 
Then the model recovers them using knowledge from context and the other modality like in Eq.~(\ref{msm_forward}). 
We denote outputs of masked tokens as $I_{Re,t}$ or $T_{Re,t}$, and $I_{Co,t}$ or $T_{Co,t}$ are output features when the corresponding tokens are not masked. 
The contrastive learning loss on masked tokens is conducted as follows: 
\begin{equation}
\begin{aligned}
\operatorname{\textbf{NCE}}_V^t=-\frac{1}{N_V^t}\sum_{i=1}^{N_V^t}\operatorname{log}\frac{\operatorname{exp}(s(I_{Re,t}^i,I_{Co,t}^i)/\tau)}{\sum_{n=1}^{N_V^t}\operatorname{exp}(s(I_{Re,t}^i,I_{Co,t}^n)/\tau)}\, , \\
\operatorname{\textbf{NCE}}_L^t=-\frac{1}{N_L^t}\sum_{i=1}^{N_L^t}\operatorname{log}\frac{\operatorname{exp}(s(T_{Re,t}^i,T_{Co,t}^i)/\tau)}{\sum_{n=1}^{N_L^t}\operatorname{exp}(s(T_{Re,t}^i,T_{Co,t}^n)/\tau)}\, ,
\end{aligned}
\label{eq:token conloss}
\end{equation}
where $N_V^t$ and $N_L^t$ refer to the number of masked tokens of images and texts in a batch, respectively. 
The recovered target of a masked token is the feature in its unmasked case, and features of other tokens serve as negative samples.  
The local semantic  completion learning loss is defined as: 
\begin{equation}
\mathcal{L}_{MLTC}=\operatorname{\textbf{NCE}}_V^t+\operatorname{\textbf{NCE}}_L^t.
\label{eq:scl_local}
\end{equation}

The overall training objective of our model is:
\begin{equation}
\mathcal{L}=\mathcal{L}_{CL}+\mathcal{L}_{ITM}+\mathcal{L}_{MLM}+\mathcal{L}_{MGSC}+\mathcal{L}_{MLTC}.
\label{eq:all loss} 
\end{equation}

\subsection{Model Architecture}
\label{model architecture}
Our model utilizes the vision encoder and text encoder to learn uni-modal representations and the fusion encoder to conduct cross-modal interactions, respectively. 
In the following, we will introduce each component in detail.

\subsubsection{Vision Encoder}
\noindent\textbf{Input. }The vision encoder takes visual data (a video or image) $I\in \mathcal{R}^{M\times3\times H \times W}$ containing $M$ frame(s) of resolution $H \times W$ as input, and when $I$ is an image, $M=1$. The visual data $I$ is first split into $M \times N$ patches $x\in \mathcal{R}^{M\times 3\times N \times P \times P}$, where $P \times P$ is the size of patches and $N=HW/P^2$. Then, the patches $x$ are transformed into $M$ sequences of vision tokens $\boldsymbol{V}=\{\boldsymbol{v}_i\}_{i=1}^M \in \mathcal R^{M\times N \times D}$, where $\boldsymbol{v}_i \in  \mathcal R^{N \times D}$ denotes the sequence of tokens for the $i^{th}$ frame in the visual data and $D$ denotes the dimension of vision tokens. Next, a learnable [CLS] token is concatenated to every token sequence $\boldsymbol{v}_i$, and we obtain $\boldsymbol{V}=\{{\boldsymbol{v}_i}\}_{i=1}^M \in \mathcal R^{M\times (N+1) \times D}$. Finally, the tokens $\boldsymbol{V}$ are summed with learnable spatial positional embeddings  $\boldsymbol{E}^s \in \mathcal{R}^{(N+1) \times D}$ and temporal positional embeddings $\boldsymbol{E}^t \in \mathcal{R}^{M \times D}$:
\begin{equation}
	\begin{aligned}
		\boldsymbol{g}^0_{ij} = \boldsymbol{v}_{ij} + \boldsymbol{E}^t_i + \boldsymbol{E}^s_j,
	\end{aligned}
	\label{f1}
\end{equation}
where all patches in the same spatial location of different frames are given the same spatial positional embedding $\boldsymbol{E}^s_j$, and all patches in the same frame share the same temporal positional embedding $\boldsymbol{E}^t_i$. 
\begin{figure}[]
	\centering
	\includegraphics[width=\linewidth]{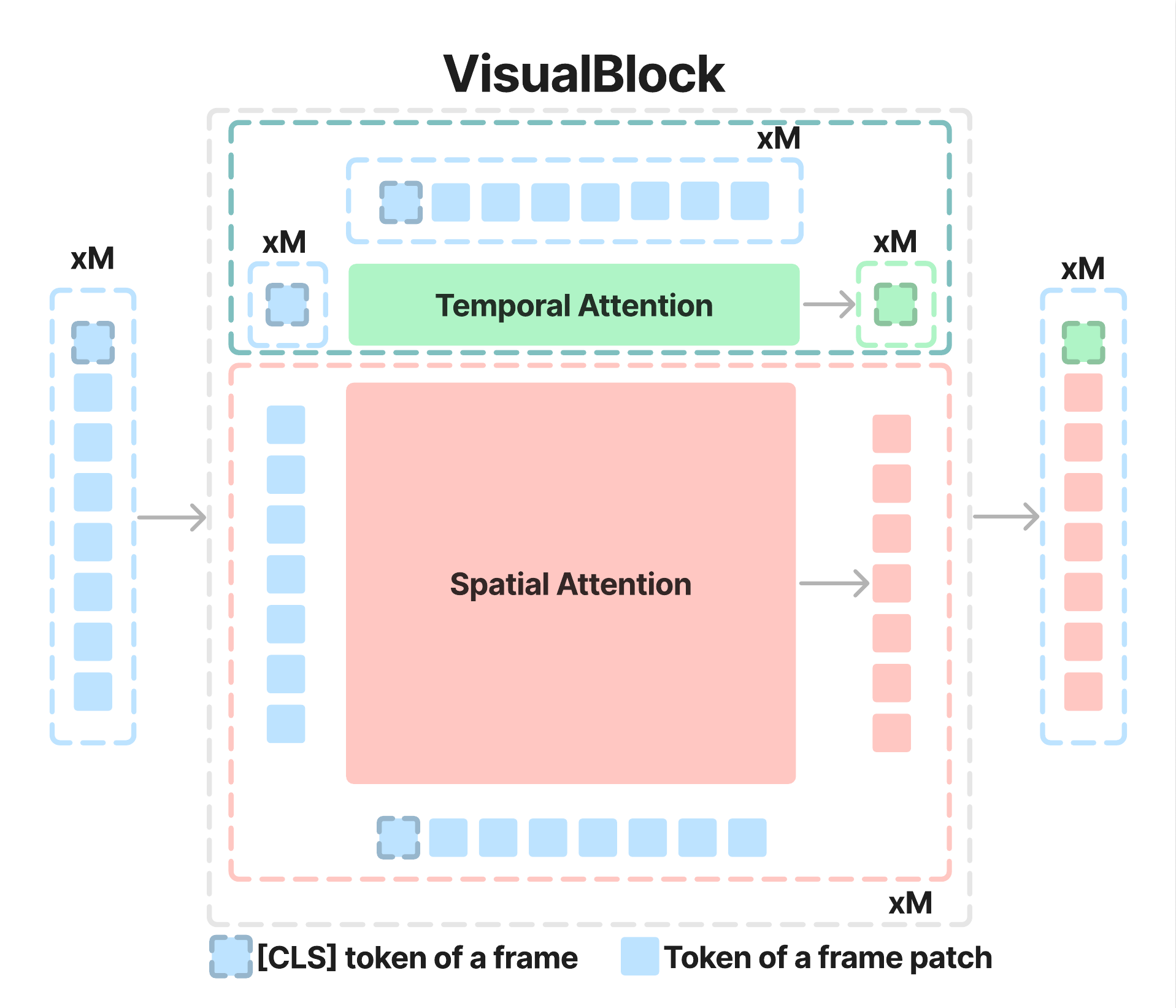}
	\centering
	\caption{The architecture of a VisualBlock.}
	\label{fig_visual_encoder}
 \vspace{-0.3cm}
\end{figure}

\noindent\textbf{VisualBlock. }
The pre-processed vision tokens $\boldsymbol{G}^{0}=\{{\boldsymbol{g}^0_i}\}_{i=1}^M \in \mathcal R^{M\times (N+1) \times D}$ are fed into the vision encoder which can process image and video data. The vision encoder is a modified ViT~\cite{DBLP:conf/iclr/DosovitskiyB0WZ21/vit}, containing a stack of VisualBlocks. 

The detail of each VisualBlock is shown in Fig. \ref{fig_visual_encoder}.
Specifically, each VisualBlock will perform temporal attention to exploit the global temporal information of the visual data and conduct the spatial attention to capture sufficient local spatial semantic information. 
For temporal attention, we perform multi-head attention for the [CLS] tokens $\{\boldsymbol{g}_{i0}^{l-1}\}_{i=1}^M$ of all frames through attending to all $M\times (N+1)$ tokens to produce [CLS] tokens $\{\boldsymbol{g}_{i0}^{l}\}_{i=1}^M$. 
Spatial attention is the multi-head attention within each frame. 
Taking the $i^{th}$ frame as an example, we use $\{\boldsymbol{g}_{ij} ^{l-1}\}_{j=1}^N$ without [CLS] token as queries and all the $N+1$ tokens in the frame as keys and values to conduct attention and obtain the output tokens $\{\boldsymbol{g}_{ij} ^{l}\}_{j=1}^N$. 
After the temporal attention and spatial attention are conducted, we concatenate the $M$ [CLS] tokens $\{\boldsymbol{g}_{i0}^{l}\}_{i=1}^M$ with the $M \times N$ tokens $\{\{\boldsymbol{g}_{ij} ^{l}\}_{j=1}^N\}_{i=1}^M$ of frame patches as the output of the VisualBlock, denoted as $\boldsymbol{G}^{l}=\{{\boldsymbol{g}^{l}_i}\}_{i=1}^M$. 

Moreover, after being processed by all the VisualBlocks, the final vision features $\boldsymbol{G}=\{{\boldsymbol{g}_i}\}_{i=1}^M \in \mathcal R^{M\times (N+1) \times D}$ can be obtained.

\subsubsection{Text Encoder}
Given the input text $T$, we first tokenize it into word embeddings $\{\boldsymbol{t}_i\}_{i=1}^K$, where $K$ is the total number of words. Then, the text encoder, which consists of a stack of bidirectional transformers \cite{DBLP:conf/nips/VaswaniSPUJGKP17/transformer},
maps $\{\boldsymbol{t}_i\}_{i=1}^K$ into token features $\boldsymbol{W}=\{\boldsymbol{w}_i\}_{i=1}^K$ by modeling the contextual relationships.

\begin{figure}[tp]
\centering
    \begin{subfigure}{0.35\textwidth}
        \centering
        \includegraphics[width=\textwidth]{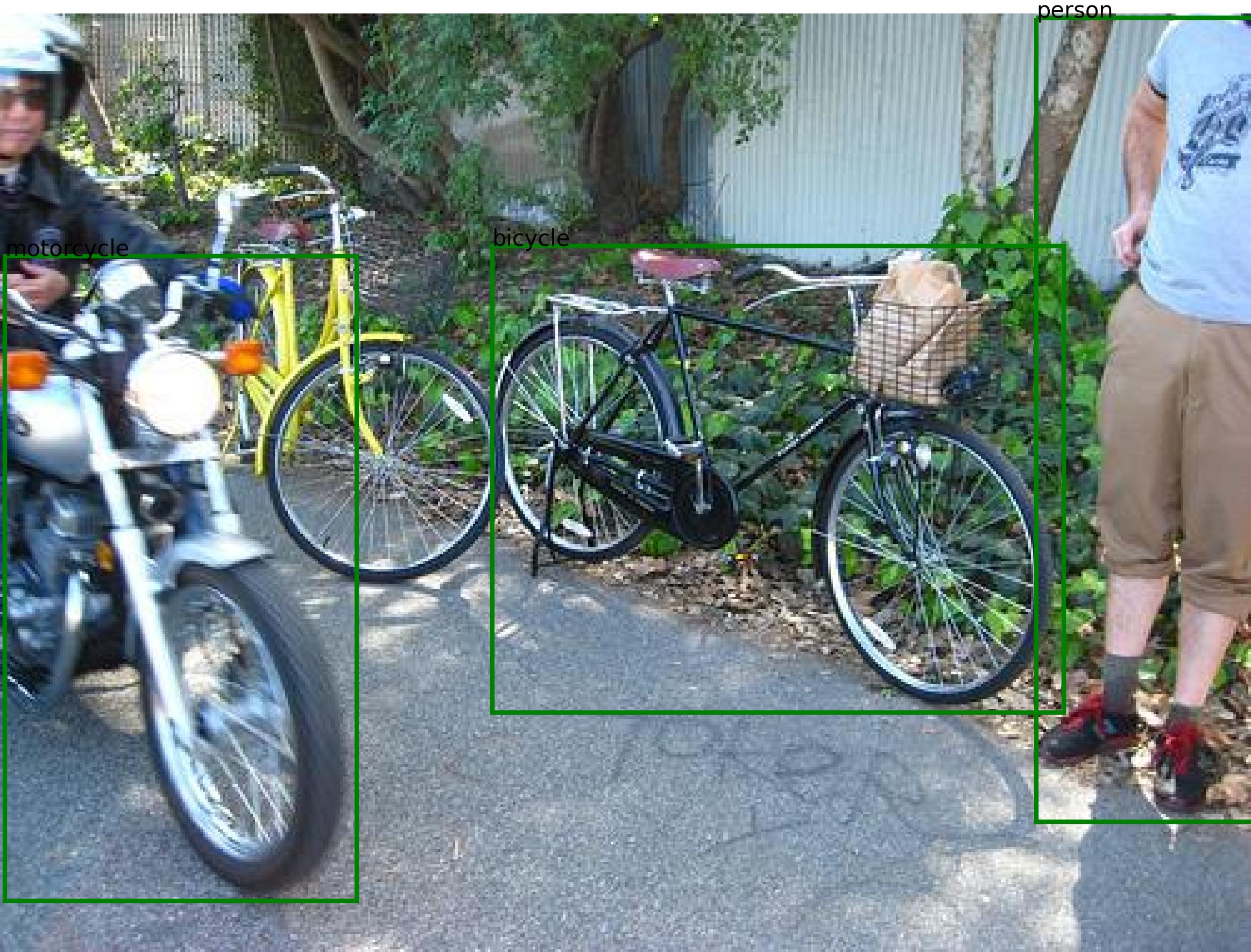}
        \caption{bounding box-level}
  \end{subfigure}
  \begin{subfigure}{0.35\textwidth}
    \centering
    \includegraphics[width=\textwidth]{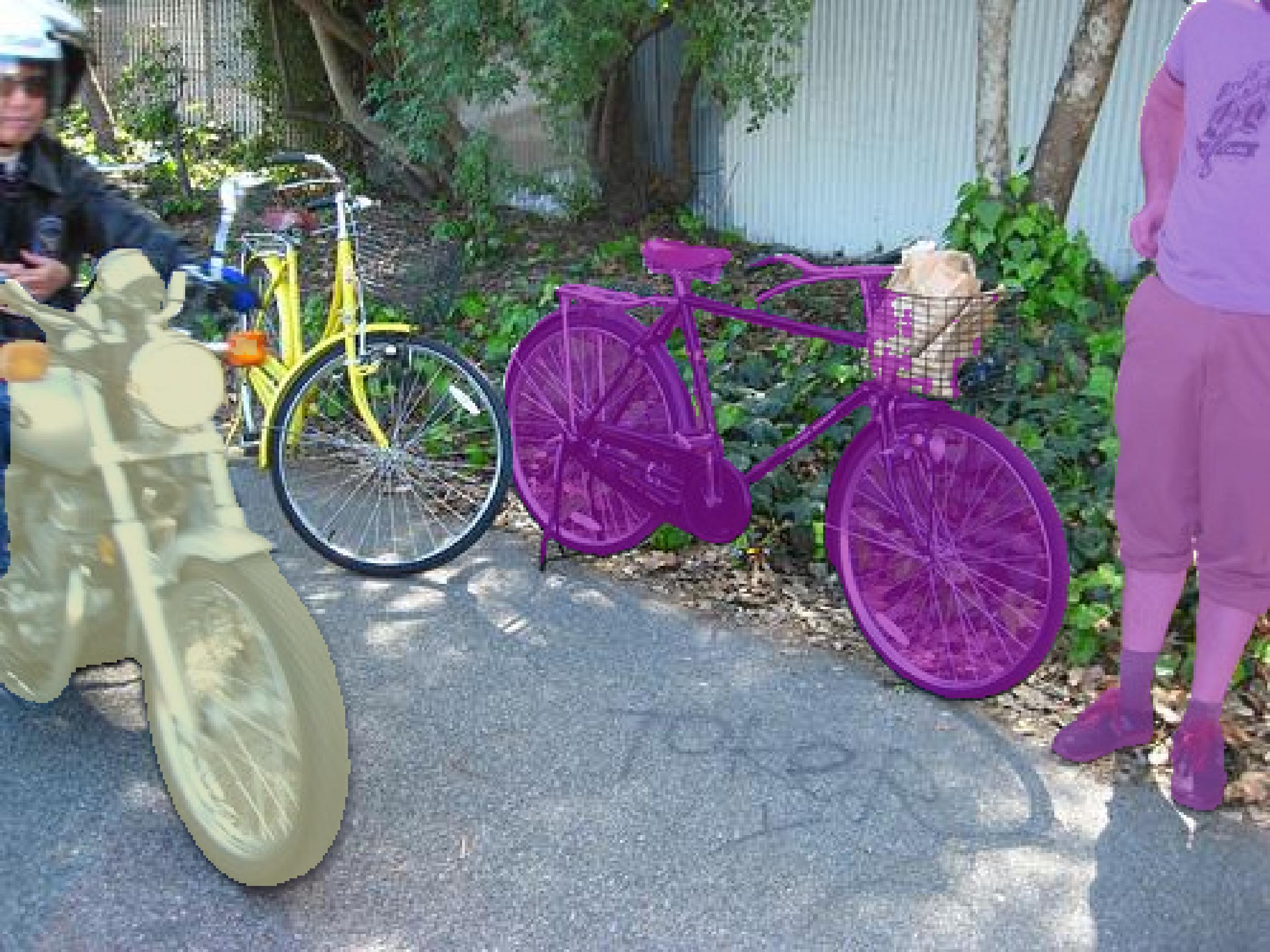}
    \caption{pixel-level}
  \end{subfigure}
  \caption{An example in our validation benchmark, and the corresponding text is `A person standing by a bicycle as a motorcycle drives by'.}
  \label{fig:example}
  \vspace{-0.3cm}
\end{figure}

\subsubsection{Fusion Encoder}
Similar to \cite{Dou_2022_CVPR/meter,DBLP:conf/emnlp/TanB19/lxmert}, we adopt a two-stream architecture for the fusion encoder, each layer of which consists of two modality-specific self-attention blocks and two cross-attention blocks. Specifically, taking the vision features as an example, following intra-modal interactions in the visual self-attention block, we conduct cross-modal interactions in the language-to-vision cross-attention block, which takes the vision tokens $\boldsymbol{G}=\{{\boldsymbol{g}_i}\}_{i=1}^M$ as queries and the text tokens $\boldsymbol{W}=\{\boldsymbol{w}_i\}_{i=1}^K$ as keys and values. The text features are conducted with similar operations. In the end, we use the mean pooling of all the frame [CLS] tokens yielded by the fusion encoder as the global representation for visual data and the [CLS] token of text as the global representation for text data.

\subsection{ALIGN-BENCH}
We argue that by adopting the proposed GLSCL as a pre-training task, the model can generate representative features by attending to regions in the images that contain abundant semantic information. 
To demonstrate this claim, we manually annotate a validation benchmark to conduct quantitative experiments on cross-modal alignment. 
Specifically, we first randomly selected 1500 image-text pairs from the MS COCO validation dataset. 
For each text-image pair, based on the corresponding text, we annotate the semantic regions in the image using both pixel-level masks and bounding box-level masks.
As shown in Fig. \ref{fig:example}, for each mask, we align each mask with its corresponding word in the text.

To evaluate cross-modal global-local alignment, we extract cross-attention maps of text [CLS] from the last layer of the fusion encoder. 
Then, we employ max-pooling to the attention maps of 12 heads, producing an attention map of text [CLS] on the image features, denoted as $Att_{cls}$. 
Meanwhile, the annotated box masks of different objects in the text are combined into a whole 0-1 mask, and so are the pixel masks. 
We resize the masks to image latent size and gain $Mask_{box,cls}$ and $Mask_{pixel,cls}$. 
Finally, we calculate the global-local alignment score as follows:
\begin{equation}
\begin{aligned}
\operatorname{\textbf{G-L}}_{box}=\frac{\operatorname{sum}(Mask_{box,cls}\cdot Att_{cls})}{\operatorname{sum}(Att_{cls})}\, , \\
\operatorname{\textbf{G-L}}_{pixel}=\frac{\operatorname{sum}(Mask_{pixel,cls}\cdot Att_{cls})}{\operatorname{sum}(Att_{cls})}\, .
\end{aligned}
\label{eq:global-local score}
\end{equation}
The cross-modal local-local alignment score of a text-image pair is the mean value of all object tokens' scores in the text. 
For each token, we obtain its attention map $Att_{token}$ similarly to [CLS]. 
The box mask and pixel mask of the token are resized to generate $Mask_{box,token}$ and $Mask_{pixel,token}$. 
Then, the local-local score is calculated as follows:
\begin{equation}
\begin{aligned}
\operatorname{\textbf{L-L}}_{box}=\frac{\operatorname{sum}(Mask_{box,token}\cdot Att_{token})}{\operatorname{sum}(Att_{token})}\, , \\
\operatorname{\textbf{L-L}}_{pixel}=\frac{\operatorname{sum}(Mask_{pixel,token}\cdot Att_{token})}{\operatorname{sum}(Att_{token})}\, .
\end{aligned}
\label{eq:local-local score}
\end{equation}

\section{Experiments}

\subsection{Implementation Details}
\subsubsection{Pre-training Settings}

Following a recent line of works, we use COCO~\cite{DBLP:conf/eccv/LinMBHPRDZ14/mscoco}, Visual Genome (VG)~\cite{DBLP:journals/ijcv/KrishnaZGJHKCKL17/vg}, Conceptual Captions (CC3M)~\cite{DBLP:conf/acl/SoricutDSG18/cc3m}, and SBU Captions~\cite{DBLP:conf/nips/OrdonezKB11/sbu} for image-text pre-training, which contain 4M images in total. 
Then, the pre-trained model is applied to image-text downstream tasks and the initialization for video-text pre-training. 
For the following video-text pre-training, we utilize WebVid~\cite{bain2021frozen} with 2.5M videos as the pre-training corpus. 
In the image-text pre-training phase, we train the model for 100k steps using a batch size of 4096 on 64 NVIDIA A100 GPUs. 
We adopt the AdamW optimizer with a weight decay of 0.01. 
The learning rate of uni-modal encoders is warmed up from 0 to $1e-5$ in the first $10\%$ steps and then decayed linearly. 
The fusion transformer has a five times higher learning rate. 
As for the video-text pre-training, the model is trained for 10k steps with the same batch size. 
The maximal learning rate of uni-modal encoders is $5e-6$, and other settings are similar to the first phase. 

As the ablation study for the impact of language encoder and vision encoder shown in METER~\cite{Dou_2022_CVPR/meter}, when adopting RoBERTa~\cite{DBLP:journals/corr/abs-1907-11692/roberta} as the language encoder and CLIP-ViT-224/16~\cite{DBLP:conf/icml/RadfordKHRGASAM21/clip} as vision encoder, the model can achieve the
most robust and decent performance. 
Therefore, we utilize CLIP-ViT-224/16~\cite{DBLP:conf/icml/RadfordKHRGASAM21/clip} and RoBERTa~\cite{DBLP:journals/corr/abs-1907-11692/roberta} to initialize vision and language encoders following the setting of METER~\cite{Dou_2022_CVPR/meter}.
The fusion encoder consists of dual-stream cross-modal blocks of 6 layers, each with a hidden dimension of 768 and 12 heads in the multi-head attention. 
As for data pre-processing, the image size is set to $288\times 288$ for pre-training and $384\times 384$ for fine-tuning, respectively. 
RandAugment~\cite{DBLP:conf/nips/CubukZS020/randaug} is applied for data augmentation. 
We resize each video frame to $224\times 224$ and uniformly sample 4 frames as video input. 
Moreover, the maximum length of input text is 50. 
For our proposed MGSC, we adopt high mask ratios, 80$\%$ for images and 40$\%$ for texts. 
In MLTC, the mask ratio of images is set to 30$\%$ by experimental exploration, and the mask ratio of texts is the same as G-SCL. 
The temperature hyper-parameter $\tau$ in Eq.~(\ref{eq:nceloss}) is set as 0.03. 

\begin{table}[]
	\centering
	\begin{adjustbox}{max width=0.5\textwidth}
		\begin{tabular}{lcccc}
			\toprule[1pt]
			\multicolumn{1}{l|}{\multirow{2}{*}{Model}} & \multicolumn{2}{c|}{\textbf{VQA2.0}}                 & \multicolumn{2}{c}{\textbf{NLVR2}}          \\
			\multicolumn{1}{l|}{}                       & test-dev       & \multicolumn{1}{c|}{test-std}       & dev            & \multicolumn{1}{l}{test-p} \\ \midrule
			\multicolumn{5}{l}{\textit{Pre-trained with \textgreater{}10M images}}                                                                           \\ \midrule
			\multicolumn{1}{l|}{ALBEF(14M)~\cite{li2021albef}}             & 75.84          & \multicolumn{1}{c|}{76.04}          & 82.55          & 83.14                      \\
			\multicolumn{1}{l|}{SimVLM~\cite{DBLP:conf/iclr/WangYYDT022/simvlm}}                 & 77.87          & \multicolumn{1}{c|}{78.14}          & 81.72          & 81.77                      \\
			\multicolumn{1}{l|}{OFA~\cite{DBLP:conf/icml/WangYMLBLMZZY22/ofa}}                    & 78.0           & \multicolumn{1}{c|}{78.1}           & -              & -                          \\
			\multicolumn{1}{l|}{BLIP~\cite{DBLP:conf/icml/0001LXH22/blip}}                   & 78.25          & \multicolumn{1}{c|}{78.32}          & 82.15          & 82.24                      \\ \midrule
			\multicolumn{5}{l}{\textit{Pre-trained with \textless{}10M images}}                                                                              \\ \midrule
			\multicolumn{1}{l|}{Oscar~\cite{DBLP:conf/eccv/Li0LZHZWH0WCG20/oscar}}                  & 73.16          & \multicolumn{1}{c|}{73.44}          & 78.07          & 78.36                      \\
			\multicolumn{1}{l|}{UNITER~\cite{chen2020uniter}}                 & 72.70          & \multicolumn{1}{c|}{72.91}          & 77.18          & 77.85                      \\
			\multicolumn{1}{l|}{ViLT~\cite{DBLP:conf/icml/KimSK21/vilt}}                   & 71.26          & \multicolumn{1}{c|}{-}              & 75.70           & 76.13                      \\
			\multicolumn{1}{l|}{TCL~\cite{DBLP:conf/cvpr/YangDTXCCZCH22/tcl}}                    & 74.90          & \multicolumn{1}{c|}{74.92}          & 80.54          & 81.33                      \\
			\multicolumn{1}{l|}{VLMo\cite{DBLP:journals/corr/abs-2111-02358/vlmo}}                 & 76.64          & \multicolumn{1}{c|}{76.89}          & 82.77          & 83.34                      \\
			\multicolumn{1}{l|}{METER~\cite{Dou_2022_CVPR/meter}}                  & 77.68          & \multicolumn{1}{c|}{77.64}          & 82.33          & 83.05                      \\
			\multicolumn{1}{l|}{Ours (MGSC)}                   & 78.72 & \multicolumn{1}{c|}{78.78} & \textbf{83.63} & 84.27             \\
           \multicolumn{1}{l|}{Ours (GLSCL)}                   & \textbf{78.91} & \multicolumn{1}{c|}{\textbf{78.92}} & 83.33 & \textbf{84.98}      \\ \bottomrule[1pt]
		\end{tabular}
	\end{adjustbox}
	\caption{Performance comparison on VQA2.0 and NLVR2. }
	\label{table:vqanlvr2}
 
\end{table}

\begin{table}[]
	\centering
	\small
	\begin{adjustbox}{max width=0.5\textwidth}
		\begin{tabular}{lcccccc}
			\toprule[1pt]
			\multicolumn{1}{l|}{\multirow{2}{*}{Model}} & \multicolumn{6}{c}{\textbf{Flickr30K-ZS}}                                          \\
			\multicolumn{1}{c|}{}                       & IR@1  & IR@5  & \multicolumn{1}{l}{IR@10} & TR@1 & TR@5 & \multicolumn{1}{l}{TR@10} \\ \midrule
			\multicolumn{7}{l}{\textit{Evaluate pre-trained models directly}}                                                                 \\ \midrule
			\multicolumn{1}{l|}{UNITER~\cite{chen2020uniter}}           & 66.16 & 88.40  & 92.94                     & 80.70 & 95.70 & 98.00                     \\
			\multicolumn{1}{l|}{ViLT~\cite{DBLP:conf/icml/KimSK21/vilt}}                   & 55.0    & 82.5  & 89.8                      & 73.2 & 93.6 & 96.5                      \\
			\multicolumn{1}{l|}{ALIGN~\cite{DBLP:conf/icml/JiaYXCPPLSLD21/align}}                 & 75.70    & 93.80  & 96.80                      & 88.60 & 98.70 & 99.70                      \\
			\multicolumn{1}{l|}{METER~\cite{Dou_2022_CVPR/meter}}         & 79.60 & 94.96          & 97.28                     & 90.90          & 98.30          & 99.50                      \\
			\multicolumn{1}{l|}{Ours (MGSC)}                   & 79.74         & 95.46 &  97.86            & 91.70 & 99.30 & \textbf{99.90}        \\ \multicolumn{1}{l|}{Ours (GLSCL)}                   & \textbf{79.96}         & \textbf{95.50} & \textbf{98.18}            & \textbf{92.70} & \textbf{99.40} & 99.80      \\ \midrule
			\multicolumn{7}{l}{\textit{Evaluate models fine-tuned on COCO}}                                                                    \\ \midrule
			\multicolumn{1}{l|}{ALBEF~\cite{li2021albef}}            & 76.8  & 93.7  & 96.7                      & 90.5 & 98.8 & 99.7                      \\
			\multicolumn{1}{l|}{ALBEF\protect\footnotemark[1]~\cite{li2021albef}}           & \textbf{82.8}  & 96.3  & 98.1                      & 94.1 & 99.5 & 99.7                      \\
			\multicolumn{1}{l|}{TCL~\cite{DBLP:conf/cvpr/YangDTXCCZCH22/tcl}}                    & 79.6  & 95.1  & 97.4                      & 93.0   & 99.1 & 99.6                      \\
			\multicolumn{1}{l|}{Ours (MGSC)}                   & 81.74         & \textbf{96.72} & \textbf{98.54}            & \textbf{94.80} & 99.60 & \textbf{100.00}                      \\
            \multicolumn{1}{l|}{Ours (GLSCL)}                  &     {82.28}    &  96.46  &     \textbf{98.54}    &  94.30  &   \textbf{99.90}   &         \textbf{100.00}        \\
			\bottomrule[1pt]
		\end{tabular}
	\end{adjustbox}
	\caption{Performance comparison of zero-shot image-text retrieval on Flickr30K.}
	\label{table:f30k-zs}
   \vspace{-0.3cm}
\end{table}

\footnotetext[1]{Pre-trained on 14M images.}

\begin{table*}[]
	\centering
	\begin{adjustbox}{max width=\textwidth}
		\begin{tabular}{lcccccccccccc}
			\toprule[1pt]
			\multicolumn{1}{l|}{\multirow{2}{*}{Model}} & \multicolumn{6}{c|}{\textbf{COCO}}                                                                                                                             & \multicolumn{6}{c}{\textbf{Flickr30K}}                                                                                                                 \\
			\multicolumn{1}{l|}{}                       & IR@1           & IR@5           & \multicolumn{1}{l}{IR@10} & \multicolumn{1}{l}{TR@1} & \multicolumn{1}{l}{TR@5} & \multicolumn{1}{l|}{TR@10}          & IR@1           & IR@5           & \multicolumn{1}{l}{IR@10} & \multicolumn{1}{l}{TR@1} & \multicolumn{1}{l}{TR@5} & \multicolumn{1}{l}{TR@10} \\ \midrule
			\multicolumn{13}{l}{\textit{Pre-trained with \textgreater{} 10M images}}                                                                                               \\ \midrule
			\multicolumn{1}{l|}{ALIGN~\cite{DBLP:conf/icml/JiaYXCPPLSLD21/align}}                  & 59.9           & 83.3           & 89.8                      & 77.0                     & 93.5                     & \multicolumn{1}{c|}{96.9}           & 84.9           & 97.4           & 98.6                      & 95.3                     & 99.8                     & 100.0                       \\
			\multicolumn{1}{l|}{ALBEF(14M)~\cite{li2021albef}}             & 60.7           & 84.3           & 90.5                      & 77.6                     & 94.3                     & \multicolumn{1}{c|}{97.2}           & 85.6           & 97.5           & 98.9                      & 95.9                     & 99.8                     & 100.0                     \\ \midrule
			\multicolumn{13}{l}{\textit{Pre-trained with \textless{} 10M images}}                                                                                                                                                                   \\ \midrule
			\multicolumn{1}{l|}{UNITER~\cite{chen2020uniter}}                 & 52.93          & 79.93          & 87.95                     & 65.68                    & 88.56                    & \multicolumn{1}{c|}{93.76}          & 75.56          & 94.08          & 96.76                     & 87.30                    & 98.00                    & 99.20                     \\
			\multicolumn{1}{l|}{PixelBERT~\cite{DBLP:journals/corr/abs-2004-00849/pixel-bert}}              & 50.1           & 77.6           & 86.2                      & 63.6                     & 87.5                     & \multicolumn{1}{c|}{93.6}           & 71.5           & 92.1           & 95.8                      & 87.0                     & 98.9                     & 99.5                      \\
			\multicolumn{1}{l|}{VinVL~\cite{DBLP:conf/cvpr/ZhangLHY0WCG21/vinvl}}                  & 58.1           & 83.2           & 90.1                      & 74.6                     & 92.6                     & \multicolumn{1}{c|}{96.3}           & -              & -              & -                         & -                        & -                        & -                         \\
			\multicolumn{1}{l|}{ALBEF(4M)~\cite{li2021albef}}              & 56.80          & 81.50          & 89.20                     & 73.10                    & 91.40                    & \multicolumn{1}{c|}{96.00}          & 82.80          & 96.70          & 98.40                     & 94.30                    & 99.40                    & 99.80                     \\
			\multicolumn{1}{l|}{SOHO~\cite{huang2021seeing/soho}}                   & 50.6           & 78.0           & 86.7                      & 66.4                     & 88.2                     & \multicolumn{1}{c|}{93.8}           & 72.5           & 92.7           & 96.1                      & 86.5                     & 98.1                     & 99.3                      \\
			\multicolumn{1}{l|}{TCL~\cite{DBLP:conf/cvpr/YangDTXCCZCH22/tcl}}                    & 59.0           & 83.2           & 89.9                      & 75.6                     & 92.8                     & \multicolumn{1}{c|}{96.7}           & 84.0           & 96.7           & 98.5                      & 94.9                     & 99.5                     & 99.8                      \\
			\multicolumn{1}{l|}{METER~\cite{Dou_2022_CVPR/meter}}                  & 57.08          & 82.66          & 90.07                     & 76.16                    & 93.16                    & \multicolumn{1}{c|}{96.82}          & 82.22          & 96.34          & 98.36                     & 94.30                    & 99.60                    & 99.90                     \\
			\multicolumn{1}{l|}{Ours (MGSC)}                   & 60.14 & 84.56 & 91.45            & 77.70           & 94.10            & \multicolumn{1}{c|}{97.44} & 84.56 & 97.42 & 98.94            & \textbf{95.90}            & 99.80           & \textbf{100.00}                     \\
   \multicolumn{1}{l|}{Ours (GLSCL)}                   &   \textbf{60.37}   &   \textbf{84.89}   &     \textbf{91.74}     &    \textbf{78.40}      &    \textbf{94.58}    & \multicolumn{1}{c|}{\textbf{97.46}} & \textbf{84.84} & \textbf{97.48} & \textbf{99.16}            & \textbf{95.90}            & \textbf{100.00}            & \textbf{100.00}                     \\ \bottomrule[1pt]
		\end{tabular}
	\end{adjustbox}
	\caption{Performance comparison of fine-tuned image-text retrieval on Flickr30K and COCO datasets.}
	\label{table:f30k_coco}
 \vspace{-0.3cm}
\end{table*}
\subsubsection{Downstream Tasks}
We employ an extensive set of evaluation benchmarks on a wide variety of vision-language understanding and retrieval tasks, including visual question answering (VQA2.0~\cite{balanced_vqa_v2/vqa_v2}), visual reasoning (NLVR2~\cite{DBLP:conf/acl/SuhrZZZBA19/nlvr2}), image-text retrieval (Flickr30K~\cite{DBLP:conf/iccv/PlummerWCCHL15/f30k}, COCO~\cite{DBLP:conf/eccv/LinMBHPRDZ14/mscoco}), and video-text retrieval (MSRVTT~\cite{DBLP:conf/cvpr/XuMYR16/msrvtt}, LSMDC~\cite{DBLP:conf/cvpr/RohrbachRTS15/lsmdc}). 

\noindent\textbf{Visual Question Answering (VQA).} 
Given an image and its corresponding question, the model needs to understand visual and textual information simultaneously to predict the answer. 
We concatenate output [CLS] features of the image and question and then conduct a classification on the candidates' set of 3,129 answers. 

\noindent\textbf{Visual Reasoning (NLVR2).} 
Given a pair of images and a description, the model is expected to reason whether their relationship is consistent. 
Specifically, this task is transformed into a binary classification problem. 

\noindent\textbf{Image-Text Retrieval.} 
There are two sub-tasks: (1) using images as queries to retrieve texts (TR); (2) using texts as queries to retrieve images (IR). 
The recall ratio is employed as the evaluation metric. 
We evaluate our model on Flickr30K~\cite{DBLP:conf/iccv/PlummerWCCHL15/f30k} and COCO~\cite{DBLP:conf/eccv/LinMBHPRDZ14/mscoco}. 
Flickr30K contains 1K images and 5K texts for evaluation, and COCO includes 5K images and 25K texts. 
Generally, there are five correct captions for an image. 

\noindent\textbf{Video-Text Retrieval.} 
Similar to exiting methods~\cite{DBLP:conf/cvpr/GeGLLSQL22/mcq, fu2021violet, DBLP:journals/corr/abs-2207-07885/clover}, we focus on text-to-video recall metrics. 
Our pre-trained model is evaluated on MSRVTT~\cite{DBLP:conf/cvpr/XuMYR16/msrvtt} and LSDMC~\cite{DBLP:conf/cvpr/RohrbachRTS15/lsmdc}, which both contain 1K video-text pairs for testing.

\subsection{Evaluation Results}

\subsubsection{Image-Text Understanding}
We conduct multimodal understanding tasks on VQA2.0 and NLVR2, which require the model to exploit vision and language semantic fusion. 
As the results shown in Table~\ref{table:vqanlvr2}, our model with MGSC pre-training task achieves new state-of-the-art performance compared with previous models.
For instance, when pre-trained with fewer than 10M images, our model with MGSC outperforms METER~\cite{Dou_2022_CVPR/meter} by $+1.04$ and $+1.14$ scores on VQA2.0 test-dev and test-std, respectively. 
On NLVR2, we gain $+0.86$ and $+0.93$ score improvements over the previous SOTA model VLMo~\cite{DBLP:journals/corr/abs-2111-02358/vlmo}. 
Moreover, our model pre-trained with 4M images also surpasses some models with more than 10M images, such as SimVLM~\cite{DBLP:conf/iclr/WangYYDT022/simvlm} and BLIP~\cite{DBLP:conf/icml/0001LXH22/blip}.  These results imply that by pre-training with our proposed MGSC pre-training task, the cross-modal fusion encoder will be significantly improved to generate representative
global features with accurate global-local alignment.

Furthermore, in the final row, the inclusion of MLTC leads to improved performance. Our model with GLSCL (i.e., adopting both pre-training tasks MGSC and MLTC) achieves the best performances in the majority of cases. For example, Ours (GLSCL) outperforms Ours (MGSC) by $+0.19$ and $+0.14$ scores on VQA2.0 test-dev and test-std, respectively. These results demonstrate that it is beneficial to improve the performances of downstream tasks by adopting our proposed MLTC to learn accurate local-local alignment.

\begin{table}[]
	\centering
	\begin{adjustbox}{max width=0.5\textwidth}
		\begin{tabular}{lccc|ccc}
			\toprule[1pt]
			\multicolumn{1}{l|}{\multirow{2}{*}{Model}} & \multicolumn{3}{c|}{\textbf{MSRVTT}}                 & \multicolumn{3}{c}{\textbf{LSMDC}}          \\
			\multicolumn{1}{l|}{}                       & R@1    & R@5    & R@10             & R@1    & R@5    & R@10        \\ \midrule
			\multicolumn{7}{l}{\textit{Fine-tune}}                                                                           \\ \midrule
			\multicolumn{1}{l|}{VideoCLIP~\cite{xu2021videoclip}}             &30.9     &55.4.  &66.8        &-     &-     &-                  \\
			\multicolumn{1}{l|}{Frozen~\cite{bain2021frozen}}                & 31.0    & 59.5     &70.5         &15.0     &30.8     &39.8               \\
			\multicolumn{1}{l|}{VIOLET~\cite{fu2021violet}}                 &34.5    & 63.0    & 73.4      &16.1     &36.6     &41.2             \\
			\multicolumn{1}{l|}{ALPRO~\cite{DBLP:conf/cvpr/Li0LNH22/alpro}}                  &33.9    & 60.7    & 73.2      &-     &-     &-        \\ 
			\multicolumn{1}{l|}{MCQ~\cite{DBLP:conf/cvpr/GeGLLSQL22/mcq}}                  &37.6    & 64.8    & 75.1      &17.9     &35.4     &44.5             \\ 
			\multicolumn{1}{l|}{MILES~\cite{DBLP:conf/eccv/GeGLWWSQL22}}                 &37.7     &63.6   &73.8         &17.8     &35.6      &44.1                    \\
			\multicolumn{1}{l|}{Clover~\cite{DBLP:journals/corr/abs-2207-07885/clover}}                &38.6    & 67.4    & 76.4      &22.7     &42.0     &52.6            \\ 
			\multicolumn{1}{l|}{Ours (MGSC)}                &\textbf{43.2}   & \textbf{76.0 }   & \textbf{86.7}      &\textbf{32.8}    &\textbf{62.5}    &\textbf{72.9 }       \\ \midrule
			\multicolumn{7}{l}{\textit{Zero-shot}}                                                                              \\ \midrule
			\multicolumn{1}{l|}{VideoCLIP~\cite{xu2021videoclip}}             &10.4     &22.2   &30.0            &-     &-     &-                       \\
			\multicolumn{1}{l|}{Frozen~\cite{bain2021frozen}}                 &18.7     &39.6  &51.6             &9.3     &22.0     &30.1                  \\
			\multicolumn{1}{l|}{VIOLET~\cite{fu2021violet}}                   &25.9     &49.5   &59.7            &-     &-     &-               \\
			\multicolumn{1}{l|}{ALPRO~\cite{DBLP:conf/cvpr/Li0LNH22/alpro}}                   &24.1     &44.7  &55.4       &-     &-     &-                       \\ 
			\multicolumn{1}{l|}{MCQ~\cite{DBLP:conf/cvpr/GeGLLSQL22/mcq}}                   &26.0     &46.4   &56.4         &12.2      &25.9      &32.2                     \\ 
			\multicolumn{1}{l|}{MILES~\cite{DBLP:conf/eccv/GeGLWWSQL22}}                 &26.1     &47.2   &56.9         &11.1     &24.7      &30.6                    \\
			\multicolumn{1}{l|}{Clover~\cite{DBLP:journals/corr/abs-2207-07885/clover}}                 &25.8     &49.6   &60.1         & 13.8      &28.1      &38.3         \\        
			\multicolumn{1}{l|}{Ours (MGSC)}                &\textbf{30.9}   & \textbf{54.4}    & \textbf{65.0}     &\textbf{17.2}    &\textbf{32.4}    &\textbf{39.1}       \\
			\bottomrule[1pt]
		\end{tabular}
	\end{adjustbox}
	\caption{Performance comparison of text-to-video retrieval on MSRVTT and LSMDC.}
	\label{table:msrvtt_lsmdc}
\end{table}

\begin{table}[]
	\centering
	\begin{adjustbox}{max width=0.5\textwidth}
		\begin{tabular}{lccc|ccc}
			\toprule[1pt]
			\multicolumn{1}{l|}{\multirow{2}{*}{Model}} & \multicolumn{3}{c|}{\textbf{MSRVTT}}                 & \multicolumn{3}{c}{\textbf{LSMDC}}          \\
			\multicolumn{1}{l|}{}                       & R@1    & R@5    & R@10             & R@1    & R@5    & R@10        \\ \midrule
			\multicolumn{1}{l|}{Ours (MGSC)}             &27.2     &51.1  &60.9       &16.7     &32.9    &\textbf{39.4}                \\
			\multicolumn{1}{l|}{Ours (GLSCL)}                & \textbf{30.2}    & \textbf{52.3}     &\textbf{62.7}         &\textbf{17.3}     &\textbf{33.0}     &39.2        \\
				\bottomrule[1pt]
		\end{tabular}
	\end{adjustbox}
	\caption{Zero-shot text-to-video retrieval performance of our image-text pre-training models on MSRVTT and LSMDC.}
	\label{table:zs video retrieval}
   \vspace{-0.3cm}
\end{table}

\begin{table}[]
\begin{adjustbox}{max width=\textwidth}
\begin{tabular}{l|cc|cc}
\toprule
\multirow{2}{*}{Model} & \multicolumn{2}{c|}{\textbf{Global-Local}} & \multicolumn{2}{c}{\textbf{Local-Local}} \\
                       & box score      & pixel score      & box score     & pixel score     \\ \midrule
ALBEF \cite{li2021albef}                 & 0.377          & 0.271            & 0.330         & 0.213           \\
METER \cite{Dou_2022_CVPR/meter}                 & 0.495          & 0.357            & 0.409         & 0.266           \\
Ours (MGSC)            & 0.521          & 0.366            & 0.412         & 0.276           \\
Ours (GLSCL)           & \textbf{0.568}  & \textbf{0.413}  & \textbf{0.423} & \textbf{0.282}           \\ \bottomrule
\end{tabular}
\end{adjustbox}
\caption{Quantitative evaluation of cross-modal alignment on ALIGN-BENCH.}
\label{table:align bench}
\vspace{-0.3cm}
\end{table}

\begin{table*}[t]
	\centering
	\begin{adjustbox}{max width=\textwidth}
		\begin{tabular}{c|cccccc|c|cc}
			\toprule[1pt]
			\multirow{2}{*}{Pre-training tasks} & \multicolumn{6}{c|}{\textbf{Flickr30K-ZS}}          & \textbf{VQA2.0}   & \multicolumn{2}{c}{\textbf{NLVR2}} \\
			& IR@1  & IR@5  & IR@10 & TR@1 & TR@5 & TR@10 & test-dev & dev         & test-p      \\ \midrule
			Ours (MLM+VTM+CL+MGSC)            & \textbf{70.4} & \textbf{91.1} & \textbf{95.2} & \textbf{86.2} & \textbf{97.3} & 99.2  & 77.59    & 81.01       & 81.89       \\
			w/o MLM                             & 67.6 & 89.5 & 94.4 & 84.7 & 97.5 & 99.3  & 75.46    & 78.93       & 79.41       \\
			w/o VTM                             &   49.7    &   78.5   &  87.2    &  61.1   &  86.4  &    94.7  & 76.96    & 77.86       & 79.54       \\
			w/o CL                              & 67.8 & 90.7 & 94.9 & 82.5 & 97.3 & 99.1  & \textbf{77.66}    & \textbf{81.10}        & \textbf{82.18}       \\   
			w/o MGSC                             & 67.0 & 89.4 & 94.3 & 80.0   & 96.5 & \textbf{99.4}  & 77.31    & 80.30        & 81.46      \\
			\bottomrule[1pt]
		\end{tabular}
	\end{adjustbox}
	\caption{Ablation study of each pre-training task in our model (MGSC). Note that the model pre-trained without VTM can only conduct zero-shot retrieval with vision and text encoders without feature fusion, so the recall metrics are not comparable to the others.}
	\label{table:pre-training tasks}
 \vspace{-0.3cm}
\end{table*}

\begin{table}[]
    \centering
	\begin{adjustbox}{max width=\linewidth}
		\begin{tabular}{c|cccc|c|c}
			\toprule[1pt]
			\multirow{2}{*}{Pre-training tasks} & \multicolumn{4}{c|}{\textbf{Flickr30K-ZS}}                          & \textbf{VQA2.0}   & \textbf{NLVR2}  \\
			& IR@1                      & IR@5  & \multicolumn{1}{l}{TR@1} & TR@5 & test-dev & test-p \\ \midrule
			MLM+VTM+CL                    & 67.0                     & 89.4 & 80.0                       & 96.5 & 77.31    & 81.46  \\
			+MLSC                               & 67.0                     & 90.0 & 80.5                    & 97.1 & \textbf{77.60}     & 81.88  \\
			+MVSC                               & 69.5                    & 90.5 & 85.5                     & \textbf{98.3} & 77.50     & 81.69  \\
			+MGSC        & \textbf{70.4} & \textbf{91.1} & \textbf{86.2}  & 97.3 & 77.59    & \textbf{81.89}  \\ \bottomrule[1pt]
		\end{tabular}
	\end{adjustbox}
	\caption{Ablation study of MLSC and MVSC, which are two parts of MGSC.}
	\label{table:scl pre-training tasks}
 \vspace{-0.3cm}
\end{table}

\subsubsection{Image-Text Retrieval}
We evaluate image-text retrieval in both zero-shot and fine-tuning scenarios. 
Our model achieves substantial performance improvements on Flickr30K and COCO datasets with similar model sizes and pre-training data scales. 
The results are also competitive with models pre-trained on larger datasets, such as  ALIGN~\cite{DBLP:conf/icml/JiaYXCPPLSLD21/align} and ALBEF~\cite{li2021albef}. 
In the fine-tuning phase, the model is trained with CL and VTM losses. 
During inference, for the sake of efficiency, we first filter top-k candidates with vision and language encoders and then compute VTM scores for ranking. 

To investigate the generalization ability of our model, we conduct zero-shot experiments on the Flickr30K dataset. 
As shown in Table~\ref{table:f30k-zs}, both our models with the MGSC or the whole GLSCL pre-training tasks achieve the best performance in both settings of zero-shot retrieval on Flickr30K. 
When we evaluate with the pre-trained model directly, ours (MGSC) gains a comprehensive boost from previous methods, reaching 79.74$\%$ and 91.7$\%$ in terms of IR@1 and TR@1. 
When evaluated with the model fine-tuned on COCO, ours (MGSC) outperforms models pre-trained on datasets of similar sizes, including ALBEF~\cite{li2021albef} and TCL~\cite{DBLP:conf/cvpr/YangDTXCCZCH22/tcl}.  
These findings indicate that our proposed global semantic completion strategies are effective in enhancing the quality of generated global representations, thus further improving the downstream retrieval tasks. 
Additionally, in both settings,  ours (GLSCL) has a more impressive performance in most cases, implying the effectiveness of our proposed MLTC pre-training task. 
Note that due to fine-tuning on COCO in the second situation, the influence of pre-training tasks may decline to some extent.

For fine-tuning experiments, our model with the MGSC pre-training task surpasses previous models by a large margin, as shown in Table~\ref{table:f30k_coco}. 
TCL~\cite{DBLP:conf/cvpr/YangDTXCCZCH22/tcl} has a distinguished retrieval performance with triple contrastive learning, which is cross-modal, intra-modal, and global-local. 
Compared with TCL, our method brings $+1.14\%/+0.56\%$ IR@1 boost and $+2.10\%/+1.00\%$ TR@1 on COCO and Flickr30K, respectively. 
It is worth noting that our model also has higher scores than ALIGN~\cite{DBLP:conf/icml/JiaYXCPPLSLD21/align} with 1.8B image-text pairs pre-trained. 
Thanks to global semantic completion learning, the global features capture more cross-modal information, leading to an encouraging performance on retrieval. 
Furthermore, ours (GLSCL) outperforms ours (MGSC) on both COCO and Flickr30K, demonstrating the local-local alignment can reduce the semantic gap between different modalities and enhance the model's understanding and matching capability on cross-modal information.

\subsubsection{Video-Text Retrieval}
Due to our adaptable vision encoder, the image-text pre-trained model can be readily transferred to video-text pre-training. 
We evaluate text-to-video retrieval on two popular datasets, MSRVTT and LSMDC, to prove the performance of the video pre-training model. 
Table~\ref{table:msrvtt_lsmdc} summarizes results under both fine-tuning and zero-shot settings. 
In the fine-tuning situation, compared with the previous SOTA model, our model with MGSC achieves notable performance improvements with $+4.6\%$ and $+10.1\%$ in R@1 on MSRVTT and LSMDC. 
When doing zero-shot retrieval, MGSC also gains remarkable improvements over the existing methods with $+4.8\%$ and $+3.4\%$ in R@1 on MSRVTT and LSMDC, respectively. 
These results demonstrate that the knowledge of our model learned from image-text data can be used to improve the performance of video-text retrieval tasks.

Due to limitations in computational resources, we did not pre-train our model with the MLTC task on the WebVid dataset. 
However, in order to demonstrate the generalization of MLTC for video retrieval tasks, we conduct zero-shot text-to-video retrieval experiments using both image-text pre-training models with and without MLTC.  
As shown in Table \ref{table:zs video retrieval}, ours (GLSCL) outperforms ours (MGSC) by 3.0 and 0.6 points in terms of R@1 metric on MSRVTT and LSMDC, respectively. 
These results once again highlight the effectiveness of our proposed MLTC pre-training task.

\subsubsection{Cross-modal Alignment Evaluation}
As shown in Table~\ref{table:align bench}, we conduct quantitative comparisons of cross-modal alignment on our proposed ALIGN-BENCH. 
We compare our model with ALBEF~\cite{li2021albef} and METER~\cite{Dou_2022_CVPR/meter}, which have a similar architecture so that we can utilize cross-attention maps of fusion encoders to compute alignment scores. 
Our model with MGSC outperforms the two former models on both global-local and local-local evaluation metrics. 
Furthermore, our model with GLSCL achieves the best performance. 
These results demonstrate that both the proposed MGSC and MLTC can enhance the semantic alignment capability of the pre-trained model, thus improving its performance on the downstream tasks.

\subsection{Ablation Studies}

We conduct empirical ablation experiments on pre-training tasks and the vision encoder. 
Since pre-training is time-consuming, we use COCO and VG as pre-training datasets, which is also a common setting in previous works~\cite{DBLP:conf/emnlp/TanB19/lxmert, DBLP:conf/acl/XuYLBHXH20/e2e-vlp, huang2021seeing/soho}.

\begin{table}[]
\centering
	\begin{adjustbox}{max width=\textwidth}
		\begin{tabular}{cl|cccc|c}
			\toprule[1pt]
			\multicolumn{2}{c|}{mask ratio} & \multicolumn{4}{c|}{\textbf{Flickr30K-ZS}}                              & \textbf{VQA2.0}         \\
			image          & text           & IR@1           & IR@5           & TR@1          & TR@5          & test-dev       \\ \midrule
			0.7            & 0.4            & 68.8          & 90.6         & 84.0            & \textbf{97.7} & 77.51          \\
			0.8            & 0.3            & 69.5          & 90.5         & 83.8          & 97.6          & \textbf{77.63} \\
			0.8   & 0.4   & \textbf{70.4} & \textbf{91.1} & \textbf{86.2} & 97.3          & 77.59          \\
			0.8            & 0.5            & 68.8           & 90.5         & 81.9           & 97.4           & \textbf{77.63} \\
			0.9            & 0.4            & 68.6           & 90.6           & 82.8          & 97.3           & 77.57          \\ 
			\bottomrule[1pt]
		\end{tabular}
	\end{adjustbox}
	\caption{Effect of different image and text mask ratios.}
	\label{table:mask ratio}
 \vspace{-0.3cm}
\end{table}

\begin{table*}[]
\centering
\begin{adjustbox}{max width=0.7\textwidth}
\begin{tabular}{c|cccccc|c}
\toprule[1pt]
\multirow{2}{*}{\begin{tabular}[c]{@{}c@{}}Pretraining task\\ (image mask ratio)\end{tabular}} & \multicolumn{6}{c|}{\textbf{Flickr30K-ZS}}                                                    & \textbf{VQA2.0} \\
& IR@1          & IR@5          & IR@10         & TR@1          & TR@5          & TR@10         & test-dev        \\ \midrule
w/o MLTC                                                                                       & 70.4          & 91.1          & 95.6         & 86.1          & \textbf{98.6}          & 99.4         & 77.98          \\
MLTC (0.1)                                                                                     & 70.5          & 91.3          & 95.7          & 87.4          & 98.0         & 99.4          & 78.05          \\
MLTC (0.2)                                                                                     & 71.0          & 91.3          & 95.4          & 86.3          & 97.8         & 99.0          & 78.14          \\
MLTC (0.3)                                                                                     & \textbf{71.5}          & \textbf{91.6}          & \textbf{96.0}          & \textbf{87.9}          & 98.4          & \textbf{99.5}          & 78.21           \\
MLTC (0.4)                                                                                     & 70.3          & 91.8          & 95.7          & 85.3 & 97.7          & 99.0          & 78.28           \\
MLTC (0.5)                                                                                     & 70.9 &91.2 &95.8  &84.8  &97.4  & 99.1   & \textbf{78.32}           \\
MLTC (0.6)                                                                                     & 70.8         & 91.7         & 95.5          & 86.2          & 97.7          & 99.3          &   78.21      \\  \bottomrule[1pt]
\end{tabular}
\end{adjustbox}
\caption{Ablation study of MLTC.}
\vspace{-0.3cm}
\label{table:MLTC}
\end{table*}

\begin{table}[]
    \centering
	\begin{adjustbox}{max width=\linewidth}
		\begin{tabular}{c|cccc|c}
			\toprule[1pt]
			\multirow{2}{*}{Pre-training tasks} & \multicolumn{4}{c|}{\textbf{Flickr30K-ZS}}                          & \textbf{VQA2.0}    \\
			& IR@1                      & IR@5  & \multicolumn{1}{l}{TR@1} & TR@5 & test-dev  \\ \midrule
			VTM+CL+MGSC                               & 66.9                    & 88.9 & 78.1                    & 96.1 & \textbf{76.13}     \\
			+MLTC                               & \textbf{67.9}                    & \textbf{89.1}       & \textbf{82.1} & \textbf{97.1}  &76.07 \\ \bottomrule[1pt]
		\end{tabular}
	\end{adjustbox}
	\caption{Ablation study of MLTC without MLM.}
	\label{table:MLTC w/o MLM}
\end{table}

\begin{table}[]
    \centering
	\begin{adjustbox}{max width=\linewidth}
		\begin{tabular}{c|cccc|c}
			\toprule[1pt]
			\multirow{2}{*}{Pre-training tasks} & \multicolumn{4}{c|}{\textbf{Flickr30K-ZS}}                          & \textbf{VQA2.0}    \\
			& IR@1                      & IR@5  & \multicolumn{1}{l}{TR@1} & TR@5 & test-dev  \\ \midrule
			VTM+CL+MLM                               & 66.4                    & 89.8 & 80.5                    & 97.4 & {77.69}     \\
			+MLTC                               & \textbf{68.3}                    & \textbf{90.3}       & \textbf{84.3} & \textbf{97.7}  &\textbf{77.85} \\ \bottomrule[1pt]
		\end{tabular}
	\end{adjustbox}
	\caption{Ablation study of MLTC without MGSC.}
	\label{table:MLTC w/o MGSC}
 \vspace{-0.1cm}
\end{table}

\begin{table}[]
\centering
\begin{adjustbox}{max width=\linewidth}
\begin{tabular}{c|cccccc|c}
\toprule[1pt]
\multirow{2}{*}{\begin{tabular}[c]{@{}c@{}}Loss Type\end{tabular}} & \multicolumn{6}{c|}{\textbf{Flickr30K-ZS}}                                                    & \textbf{VQA2.0} \\
& IR@1          & IR@5          & IR@10         & TR@1          & TR@5          & TR@10         & test-dev        \\ \midrule
Ours                                                                                       & \textbf{71.5}          & \textbf{91.6}          & \textbf{96.0}          & \textbf{87.9}          & \textbf{98.4}          & \textbf{99.5}          & 78.21         \\
$L_2$ loss                                                                                    & 70.1          & 91.2          & 95.4          & 85.2          & 97.2         & 99.1          & \textbf{78.22}          \\
Cosine loss                                                                                     & 70.8         & 91.4         & 95.7          & 86.7          & 97.8          & 99.4          & 78.19          \\  \bottomrule[1pt]
\end{tabular}
\end{adjustbox}
\caption{Ablation study of loss type for MGSC.}
\label{table:MGSC_loss}
\end{table}

\begin{table}[]
\centering
\begin{adjustbox}{max width=\linewidth}
\begin{tabular}{c|cccccc|c}
\toprule[1pt]
\multirow{2}{*}{\begin{tabular}[c]{@{}c@{}}Loss Type\end{tabular}} & \multicolumn{6}{c|}{\textbf{Flickr30K-ZS}}                                                    & \textbf{VQA2.0} \\
& IR@1          & IR@5          & IR@10         & TR@1          & TR@5          & TR@10         & test-dev        \\ \midrule
Ours                                                                                    & \textbf{71.5}          & \textbf{91.6}          & \textbf{96.0}          & \textbf{87.9}          & \textbf{98.4}          & \textbf{99.5}          & \textbf{78.21}          \\
$L_2$ loss                                                                                     & 69.6          & 91.5          & 95.2          & 84.4          & 98.1         & 99.4          & 77.97          \\
Cosine loss                                                                                     & 69.9         & 91.1         & 95.4          & 84.3          & 98.0          & 99.2          & 78.01          \\  \bottomrule[1pt]
\end{tabular}
\end{adjustbox}
\caption{Ablation study of loss type for MLTC.}
\label{table:MLTC_loss}
\end{table}

\begin{table}[]
\centering
\begin{adjustbox}{max width=\linewidth}
\begin{tabular}{c|cccccc|c}
\toprule[1pt]
\multirow{2}{*}{\begin{tabular}[c]{@{}c@{}}Mask Strategy\end{tabular}} & \multicolumn{6}{c|}{\textbf{Flickr30K-ZS}}                                                    & \textbf{VQA2.0} \\
& IR@1          & IR@5          & IR@10         & TR@1          & TR@5          & TR@10         & test-dev        \\ \midrule
Random-Mask                                                                                    & \textbf{71.5}          & \textbf{91.6}          & \textbf{96.0}          & \textbf{87.9}          & \textbf{98.4}          & \textbf{99.5}          & \textbf{78.21}          \\
Block-Mask                                                                                     & 70.6          & 90.4          & 94.5         & 79.1          & 96.4         & 99.0          & 78.12          \\
N\&V-Mask                                                                                     & 70.5         & 90.6         & 94.2          & 83.9          & 97.0          & 99.2          & 78.05          \\  \bottomrule[1pt]
\end{tabular}
\end{adjustbox}
\caption{Ablation study of different mask strategies.}
\label{table:mask_strategy}
\end{table}

\begin{table}[t]
	\centering
	\begin{adjustbox}{max width=\linewidth}
		\begin{tabular}{lccc|ccc}
			\toprule[1pt]
			\multicolumn{1}{l|}{\multirow{2}{*}{Vision Encoder}} & \multicolumn{3}{c|}{\textbf{Video Retrieval}}                 & \multicolumn{3}{c}{\textbf{Text Retrieval}}          \\
			\multicolumn{1}{l|}{}                       & R@1    & R@5    & R@10             & R@1    & R@5    & R@10        \\ \midrule
			\multicolumn{1}{c|}{Mean Pooling}             &23.0      &45.6      & 56.5    &22.5   & 46.4   & 54.9                 \\
			\multicolumn{1}{c|}{Global CLS}                & 22.5    & 45.9    &  55.1         &22.4   & 45.7   & 55.5            \\
			\multicolumn{1}{c|}{Frame CLS}                 &\textbf{23.3}    & \textbf{48.3}    & \textbf{56.8}     &\textbf{24.0}   &\textbf{46.5}   &\textbf{55.7}    \\
			\bottomrule[1pt]
		\end{tabular}
	\end{adjustbox}
	\caption{Ablation study of the vision encoder on zero-shot retrieval of MSRVTT.}
	\label{table:video_ablation}
 \vspace{-0.3cm}
\end{table}

\subsubsection{Different Pre-training Tasks}
\label{sec:ablation pre tasks}

There are four pretext tasks in our previous model (MGSC), including CL, VTM, MLM, and MGSC. 
As summarized in Table~\ref{table:pre-training tasks}, we explore the impact of each task on both retrieval and understanding datasets. 
The first row shows the results of our model with all pre-training tasks, and the second to fifth rows reflect the effect of removing each task separately. 
According to the chart, we observe that the retrieval performance drops most due to the lack of MGSC when conducting retrieval with feature fusion. 
Specifically, MGSC brings $+3.38\%$ and $+6.20\%$ boost in IR@1 and TR@1 on F30K-ZS. 
The model without MLM loses $2.13\%$ in the accuracy of VQA2.0, which indicates that MLM has a great effect on multimodal understanding tasks. 
As for NLVR2, VTM has a relatively large impact. 
However, contrastive learning is only effective for retrieval in our model, which is perhaps because the other three pre-training tasks have already learned cross-modal fusion sufficiently for understanding tasks. 
Overall, comparing the first row with the fifth row, the model with MGSC makes progress on all downstream tasks, which demonstrates that the model learns more accurate cross-modal alignment to generate representative global features. 

Furthermore, MGSC comprises MVSC and MLSC, whose effects we showcase in Table~\ref{table:scl pre-training tasks}. 
According to the first three rows, either MVSC or MLSC can improve the performance of downstream tasks. 
We find that MVSC has a superior impact on retrieval tasks, which is probably because it improves the robustness of visual information understanding. 
In VQA2.0 and NLVR2, MLSC plays a more important role. 
Additionally, when combining the two sub-tasks, our model performs better in most metrics, which indicates that they are in synergy. 

\begin{figure*}[tp]
	\centering
	\includegraphics[width=0.95\textwidth]{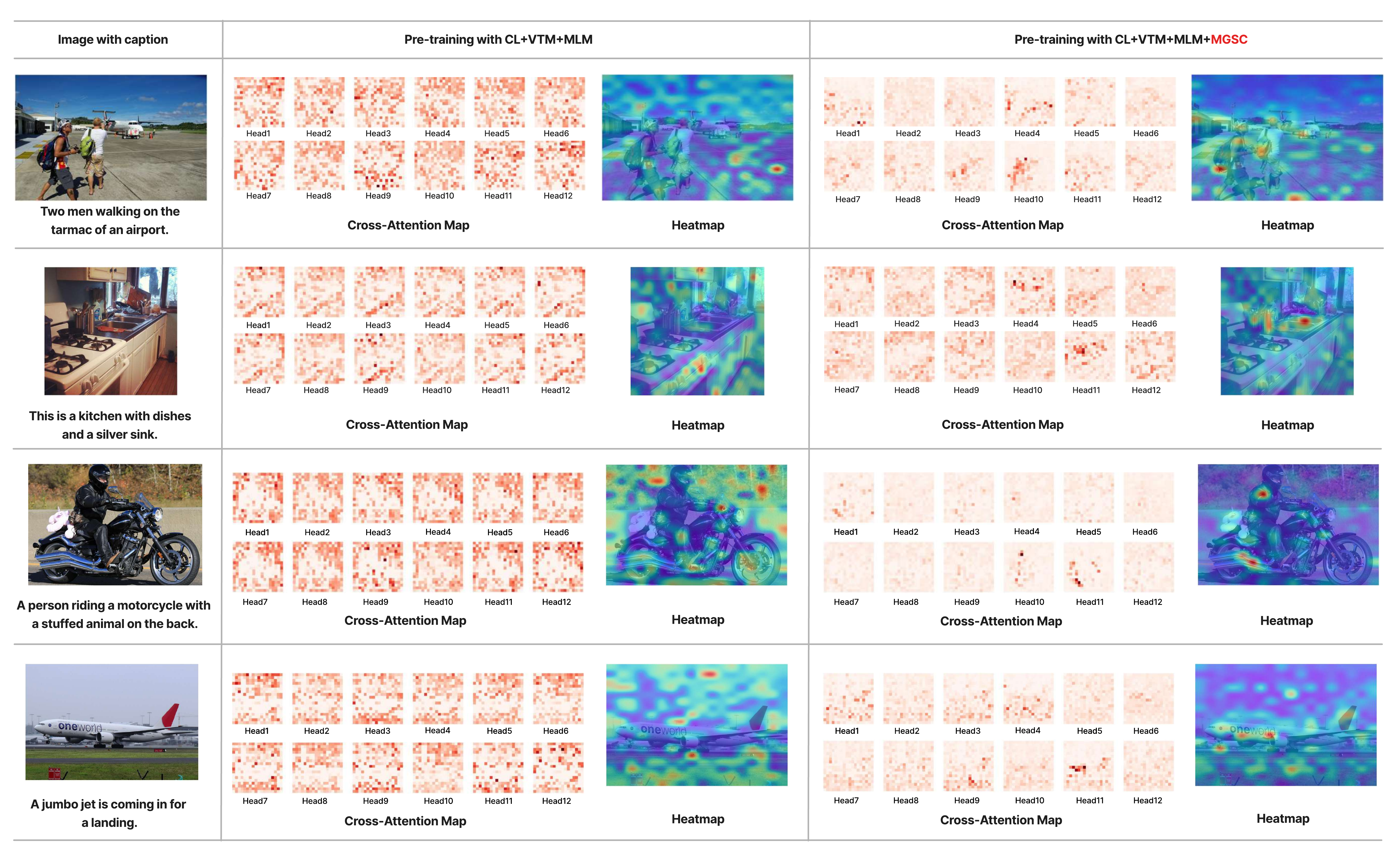}
	\caption{The cross-attention visualization of text [CLS] on the whole image for the model pre-trained with or without MGSC. The cross-modal attention maps of 12 heads are from the last layer of the fusion encoder. Then we depict heatmaps by max-pooling the attention maps.}
	\label{fig:visualization}
 \vspace{-0.3cm}
\end{figure*} 

\subsubsection{Mask Ratio in MGSC}
\label{sec:mask ratio}

As shown in Table~\ref{table:mask ratio}, we observe that the mask ratios of image and text affect downstream tasks, especially on zero-shot retrieval. 
VQA2.0 is less sensitive to the mask ratio because the model has been fine-tuned with a large amount of data. 
Considering the second to fourth rows, when the image mask ratio is fixed, the model with a text mask ratio of 0.4 has almost the best performance. 
Moreover, when the text mask ratio is set to 0.4, the results of the image mask ratio of 0.8 are the highest. 
We speculate that when the mask ratio is lower, semantic completion will rely more on intra-modal information and lack learning across modalities, leading to inferior performance. 
When the mask ratio is too high, the small number of remaining tokens can only perform very limited cross-modal interactions. 
In conclusion, we choose 0.4 and 0.8 as text and image mask ratios in MGSC, respectively. 

\subsubsection{Masked Local Token Completion} 
The ablation studies about the newly proposed MLTC and its image mask ratio are depicted in Table~\ref{table:MLTC}. 
In comparison to the ablation experiments conducted in Sec. \ref{sec:ablation pre tasks} and Sec. \ref{sec:mask ratio}, the current experiments are conducted with smaller batch sizes and more training steps due to the introduction of MLTC. 
The first row corresponds to the model pre-trained with MLM, VTM, CL and MGSC. 
The results reveal that the model with MLTC makes consistent progress on retrieval and understanding tasks. 
However, the high image mask ratio (0.8) used in MGSC is not suitable for masked vision token reconstruction. 
Thus, we investigate the impact of the image mask ratio of MLTC through ablation analysis. 
The second to seventh rows in the table illustrate that the model with a mask ratio of 0.3 performs best on zero-shot retrieval, while VQA2.0 accuracy is highest when the mask ratio is set to 0.5. 
Based on these results, we select 0.3 as the image mask ratio for MLTC.

Additionally, as the masked language modeling (MLM) task can complete local semantic information for the masked tokens of text to some extent, we further conducted an experiment without MLM to validate the effectiveness of the proposed MLTC, and the experimental results are shown in Table \ref{table:MLTC w/o MLM}.
It can be found that after adding MLTC, although there is a slight decline in VQA performance, the zero-shot retrieval results on Flickr30K show significant improvement, which directly reflects the effect of pre-training tasks. 
We infer that VQA performance relies more on MLM to learn better textual representations. 
The absence of MLM would lead to deficient textual representations,  which will impair the reconstruction goals of MLTC, thereby limiting the effectiveness of MLTC in understanding.

Therefore, we further conduct experiments without MGSC to validate the effectiveness of the proposed MLTC. As the results shown in Table \ref{table:MLTC w/o MGSC}, when the model is pre-trained with the MLTC task, it can achieve much better retrieval and visual question-answering performance.

\subsubsection{Different Types of Loss for GLSCL}
The loss functions for our proposed MGSC and MLTC pre-training tasks are constructed based on contrastive learning loss. Currently, we propose two different types of loss functions to investigate the effect of our used contrastive learning loss on the downstream tasks.  
The first loss function is based on $l_2$ loss, which is defined as follows:
\begin{equation}
\begin{aligned}
\operatorname{\textbf{Loss}}_V=\|I_{Re}^i - I_{Co}^i\|_F^2\, , \\
\operatorname{\textbf{Loss}}_L=\|T_{Re}^i - T_{Co}^i\|_F^2\, .
\end{aligned}
\label{eq:l2loss}
\end{equation}
Moreover, the second loss function is constructed by the cosine similarity:
\begin{equation}
\begin{aligned}
\operatorname{\textbf{Loss}}_V=-log(0.5(s(I_{Re}^i, I_{Co}^i)+1))\, , \\
\operatorname{\textbf{Loss}}_L=-log(0.5(s(T_{Re}^i, T_{Co}^i)+1))\, ,
\end{aligned}
\label{eq:cosloss}
\end{equation}
where $s$ refers to cosine similarity.  
To conduct the experiments, we pre-train our model by replacing Eq. (\ref{eq:nceloss}) with Eq. (\ref{eq:l2loss}) and Eq. (\ref{eq:cosloss}), respectively, while keeping the other loss functions unchanged. The corresponding results are shown in Table \ref{table:MGSC_loss}. Similarly, for MLTC, we also reconstruct masked token features with $l_2$ loss and cosine similarity loss, respectively, while leaving the other loss functions unchanged to pre-train our model. The corresponding results are presented in Table \ref{table:MLTC_loss}.

As the results shown in the two tables, by adopting Eq. (\ref{eq:nceloss}) and Eq. (\ref{eq:token conloss}) as the loss functions for our pre-trained tasks, MGSC and MLTC, our model achieves superior performance compared to the other types of losses. This can be attributed to the reason that the $l_2$ loss and cosine similarity-based loss focus solely on minimizing the distance between the masked data (or token) and its corresponding ground truth feature, whereas the contrastive learning loss takes into account other data (or tokens) and encourages the masked data (or token) to be not only similar to its ground truth feature but also different from the features of other data (or tokens). Therefore, our model is trained to learn more accurate global-local and local-local semantic alignments by adopting the contrastive learning loss, leading to better performance on downstream tasks.

\begin{figure*}[tp]
	\centering
	\includegraphics[width=0.95\textwidth]{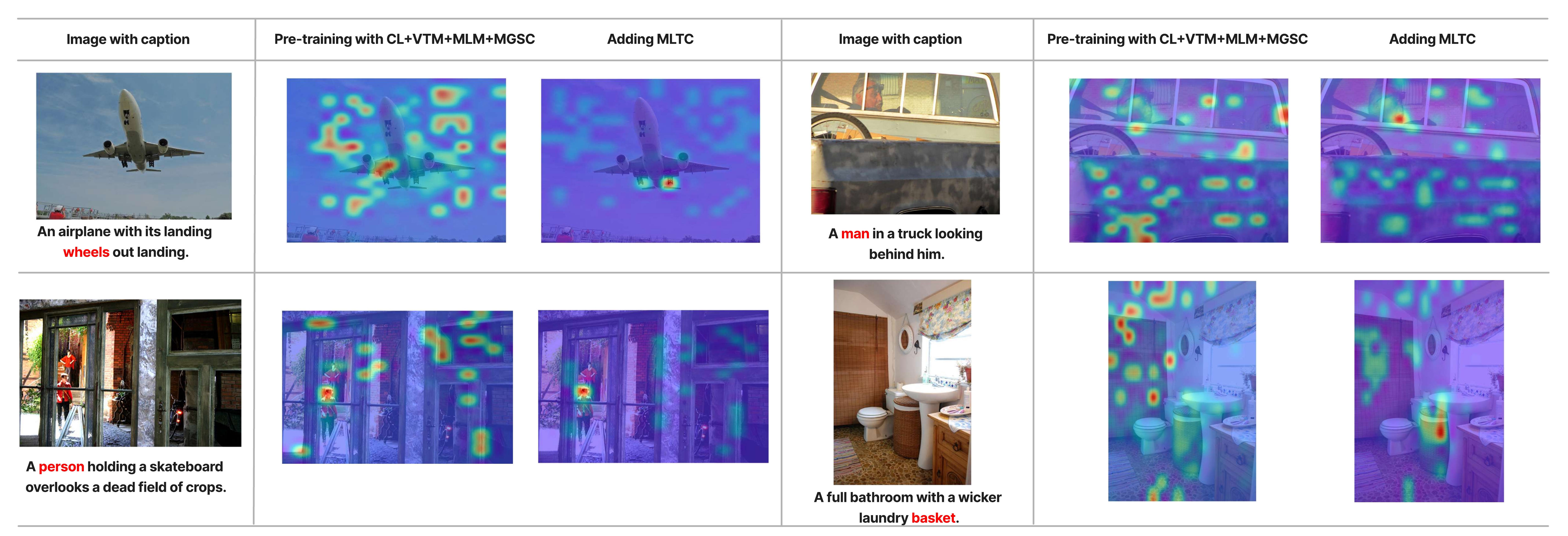}
	\caption{The cross-attention visualization of a token on the whole image for the model pre-trained with or without MLTC.}
	\label{fig:visualization_token}
 \vspace{-0.3cm}
\end{figure*}

\subsubsection{Mask Strategy}
In this section, we investigate the impact of the masking strategy on the proposed pre-training task GLSCL. 
In the original GLSCL pre-training task, we adopted a random selection approach to mask tokens in both images and texts to construct the training data. 
Then, in this experiment, we propose two additional variants: (1) The first one changes the mask strategy for image data, dubbed Block-Mask. For an image, we first randomly select a token and then use it as the center to mask the surrounding tokens. 
(2) The second one, termed N\&V-Mask, alters the masking strategy for the text data, which focuses solely on masking the nouns and verbs in the text data. 
The experimental results are shown in Table \ref{table:mask_strategy}, where Random-Mask denotes the original mask strategy used in the GLSCL pre-training task. 
Based on these results, we can observe that the Random-Mask outperforms the Block-Mask on both image-text retrieval and visual question-answering tasks. 
This superiority may be attributed to the fact that when the mask tokens are randomly selected, the model can reconstruct the mask tokens by learning features from their surrounding tokens. Moreover, the Random Mask also achieves better performance than the N\&V-Mask. 
This discrepancy may arise from the low ratio of nouns and verbs in the sentences, leading to an insufficient mask ratio that hinders the performance of the pre-trained model. 
Therefore, in other experiments, we adopt the random selection strategy to mask tokens for the proposed GLSCL pre-training task.

\subsubsection{Vision Encoder Design}
To investigate the effectiveness of our designed vision encoder in processing video data, we compare it with two other variants: (1) The first variant, termed Mean Pooling, directly treats a video as $M$ separate images and then uses the mean pooling of $M$ [CLS] tokens as the video representation. (2) The second variant, termed Global CLS, is the vision encoder proposed by MCQ~\cite{DBLP:conf/cvpr/GeGLLSQL22/mcq}. 
In this experiment, we pre-train the vision encoder and text encoder on the WebVid~\cite{bain2021frozen} dataset via contrastive learning and then conduct zero-shot cross-modal retrieval on the MSRVTT dataset. The experimental results are shown in Table \ref{table:video_ablation}, where the Frame CLS denotes our designed vision encoder. It can be found that Frame CLS achieves the best performance on both video-to-text and text-to-video retrieval tasks, which demonstrates the outstanding capability of video temporal modeling. 

\subsection{Visualization Analysis}
\label{sec:visualization}
To demonstrate that MGSC boosts cross-modal alignment for global representations, we visualize cross-attention maps between text [CLS] and the whole image in the last layer of the fusion encoder. 
We conduct max-pooling on attention maps of 12 heads to draw heatmaps, as shown in Fig.~\ref{fig:intro_visulize}(a) and Fig.~\ref{fig:visualization}. 
Compared with the baseline, the model with MGSC can recognize relevant regions more precisely. 
For example, in the first image of Fig~\ref{fig:intro_visulize}(b), the attention distribution of global text representation to the image is scattered without MGSC, while after MGSC pre-training, [CLS] pays attention to the fish, lemons, asparagus in the image. 
Taking the second image of Fig~\ref{fig:visualization} as another example, the model pre-trained with MGSC identifies the dishes and sink in the kitchen, which indicates a desirable global-local alignment ability. 

Observing attention maps of 12 heads in Fig.~\ref{fig:visualization}, we find that the attention maps without MGSC are basically the same for an image, but for the model pre-trained with MGSC, the attention maps of different heads are distinctive, which means that each head learns various information from the image. 
Overall, MGSC encourages global representations to learn cross-modal interactions, extracting useful knowledge from the other modalities. 

In Fig. \ref{fig:visualization_token}, we present visualizations of cross-attention maps for certain text tokens to illustrate the local-local alignment capability of MLTC. 
Compared to the baseline pre-trained with CL, VTM, MLM and MGSC, the regions associated with the specific words receive notable attention upon incorporating MLTC. 
For instance, in the first image, the model with MLTC pre-training successfully recognizes the plane wheels. 
Similarly, in the fifth image, the model with MLTC can find the basket in the complex bathroom environment. 
These results indicate masked local token completion enhances our model's comprehension of multimodal data.

\section{Limitation}
During the training phase, we make the assumption that the data belonging to the same visual-language pair are similar to each other, and the data from different visual-language pairs are dissimilar to each other. This assumption serves as the basis for constructing our objective function. However, there may be some noisy data pairs, where the data from the same visual-language pair exhibit dissimilarities and the data from different visual-language pairs display similarities. Neglecting these data pairs can adversely affect the performance of our model in downstream tasks. To address this, in future work, we can design an online adaptive denoising method to rectify the relationships between training data pairs, thereby enhancing the robustness of our model.


\section{Conclusion} 
In this paper, we propose a new vision-language pre-training task called Global and Local Semantic Completion Learning (GLSCL), which consists of Masked Global Semantic Completion (MGSC) and Masked Local Token Completion (MLTC). 
Different from previous pre-training tasks that reconstruct IDs or pixel values of masked local tokens, GLSCL leverages cross-modal interactions to recover global and local semantic features of masked data, promoting global-local and local-local cross-modal alignment. 
Ablation studies and visualization analysis demonstrate the effectiveness of GLSCL. 
We develop a new benchmark, ALIGN-BENCH, to quantitatively evaluate the cross-modal alignment of our model. 
Moreover, we introduce a flexible vision encoder, which adapts to image-text and video-text multimodal tasks readily. 
We conducted image-text and video-text pre-training sequentially and applied our model to various challenging downstream tasks. 
The extensive evaluations validate the great superiority of our GLSCL method.

{\small
    \bibliographystyle{IEEEtran}
    \bibliography{egbib}
}

        \begin{IEEEbiography}[{\includegraphics[width=1in,height=1.25in,clip,keepaspectratio]{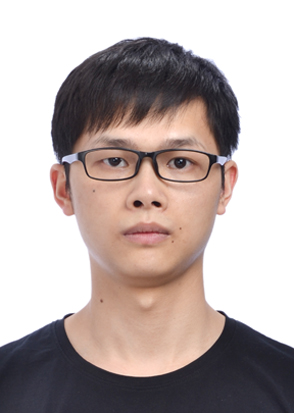}}]{Rong-Cheng Tu}
		received a bachelor's degree from Beijing Institute of Technology, China, in 2018. He is currently working toward a Ph.D. in the Department of Computer Science and Technology at the Beijing Institute of Technology, China. His research interests are in deep learning, information retrieval and learning to hash.
	\end{IEEEbiography}

        \begin{IEEEbiography}[{\includegraphics[width=1in,height=1.25in,clip,keepaspectratio]{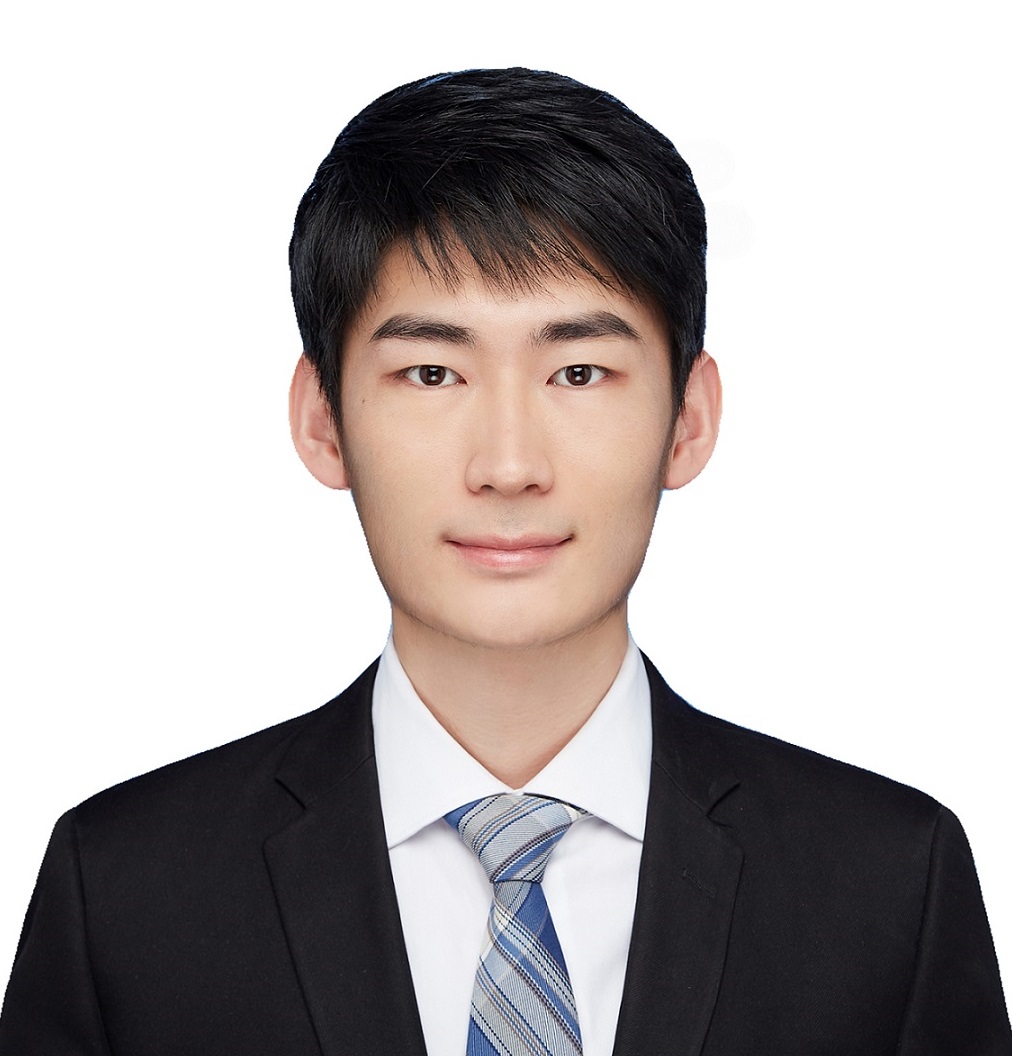}}]{Yatai Ji} 
		received the bachelor's degree from the Department of Automation, Tsinghua University, China, in 2021. He is currently a master's student in Electronic and Information Engineering at Tsinghua Shenzhen International Graduate School, Tsinghua University. His research interests are deep learning, multimodal learning, and vision-language pre-training.
	\end{IEEEbiography}

    \begin{IEEEbiography}[{\includegraphics[width=1in,height=1.25in,clip,keepaspectratio]{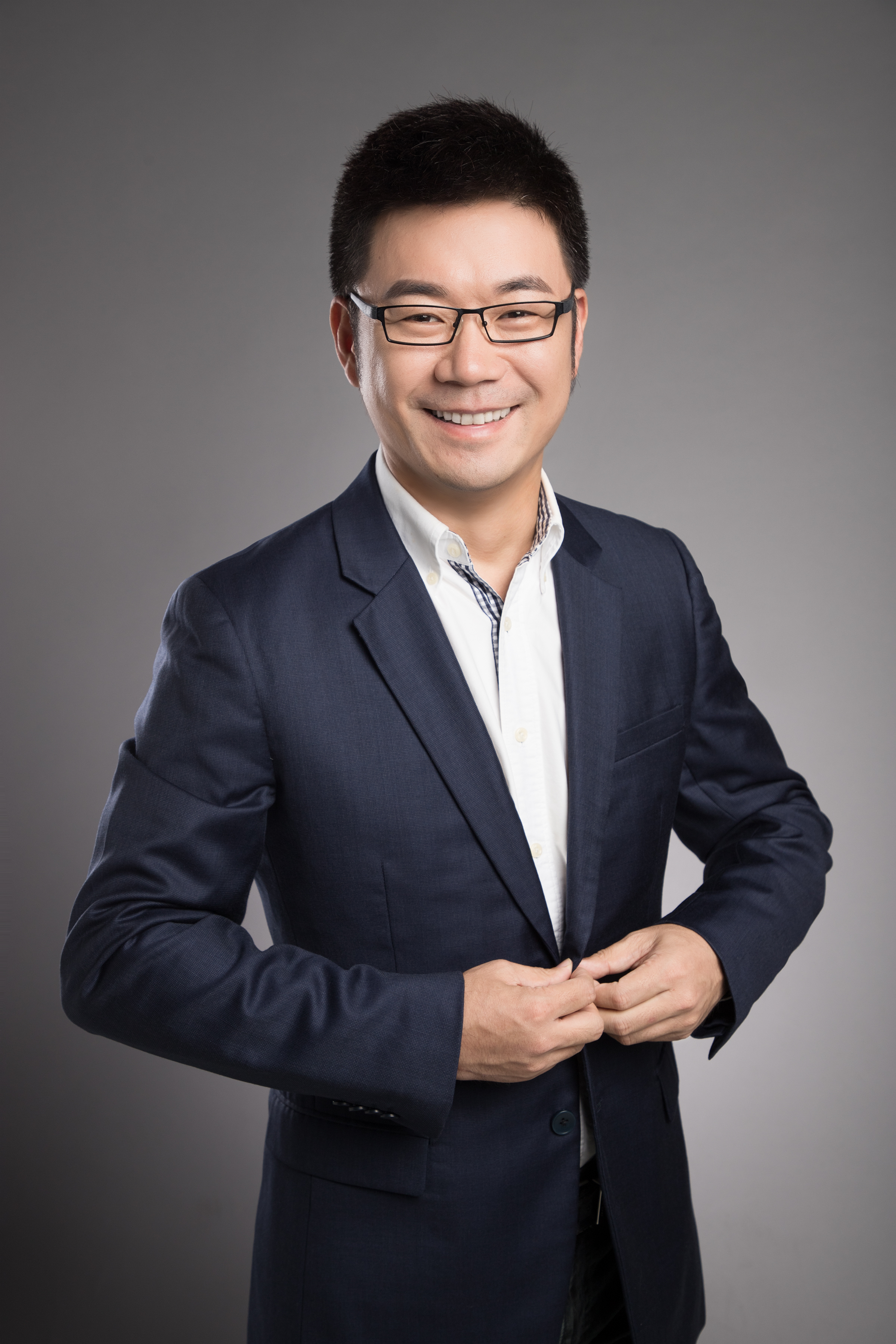}}]{Jie Jiang}
		received a Ph.D. degree from Peking University and has long been devoted to big data and distributed computing. He is well-known in China for his expertise in data science and is a member of the CCF Task Force on Big Data. He has been invited to give keynote speeches at SACC and Hadoop in China many times.
	\end{IEEEbiography}

    \begin{IEEEbiography}[{\includegraphics[width=1in,height=1.25in,clip,keepaspectratio]{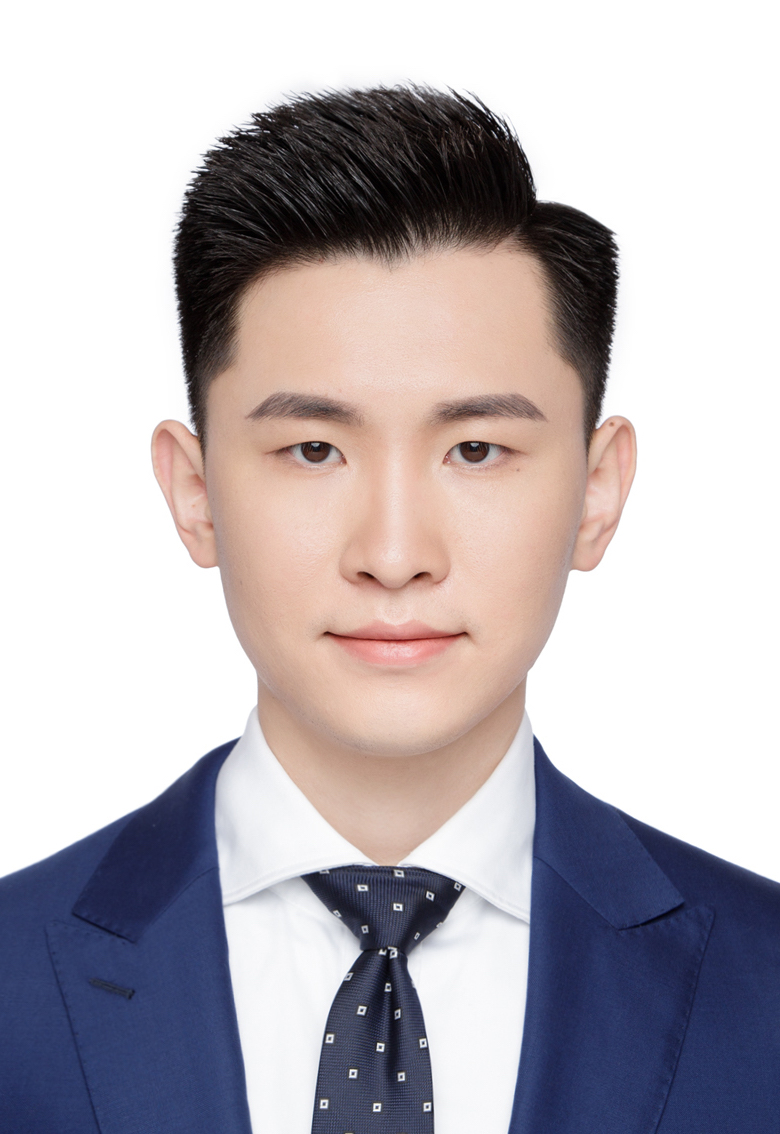}}]{Weijie Kong} 
	 received a B.S. degree in software engineering from Northeastern University, Shenyang, Liaoning, in 2017 and an M.S. degree in computer applied technology from Peking University, Beijing, in 2020. His research interest includes video understanding, content-based retrieval and multimodal representation learning.
	\end{IEEEbiography}

	\begin{IEEEbiography}[{\includegraphics[width=1in,height=1.25in,clip,keepaspectratio]{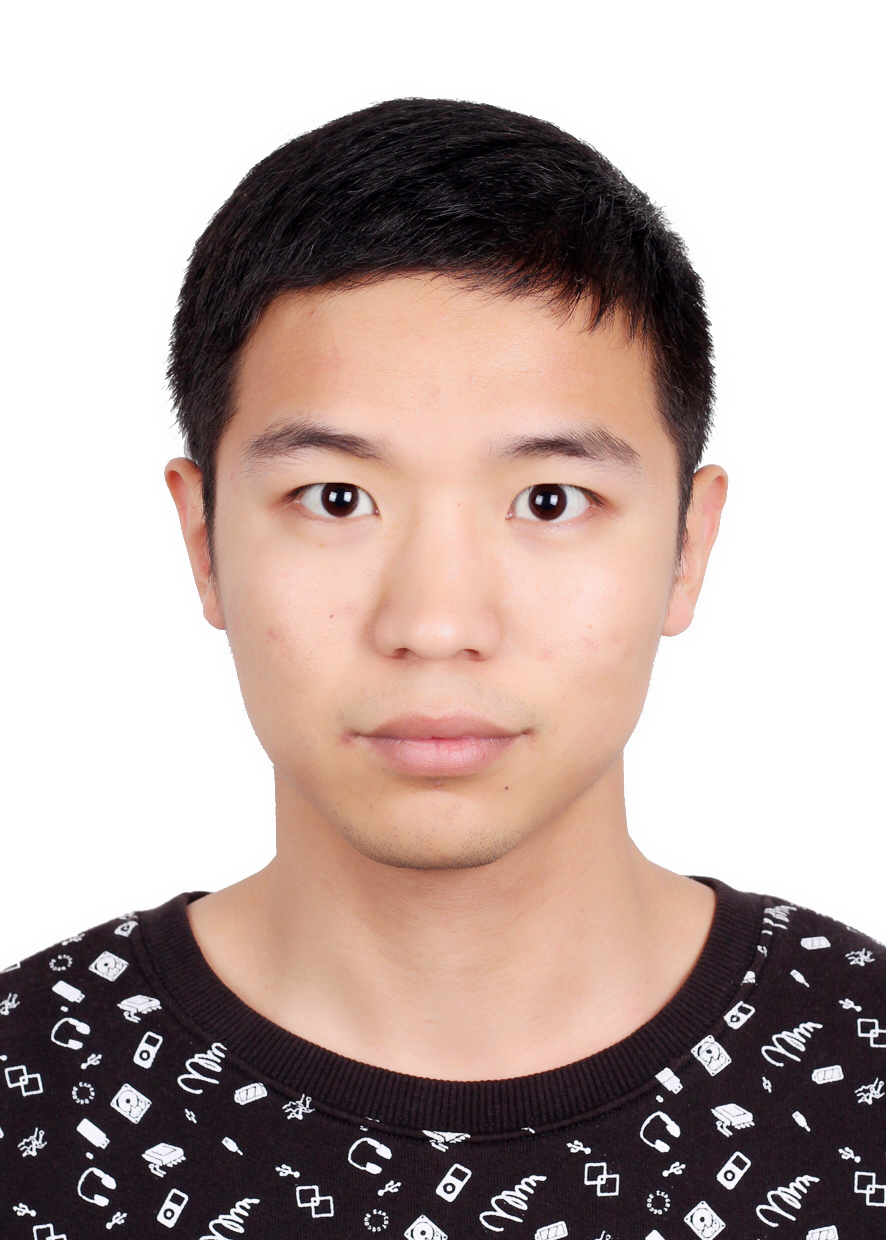}}]{Chengfei Cai}
		received a bachelor's degree  and a master's degree from Zhejiang University, China. He is currently a Senior Researcher of Ads Multimedia AI at Tencent Data Platform. His research interests are in  information retrieval, machine learning, deep learning, data mining, and computer vision.
	\end{IEEEbiography}

        \begin{IEEEbiography}[{\includegraphics[width=1in,height=1.25in,clip,keepaspectratio]{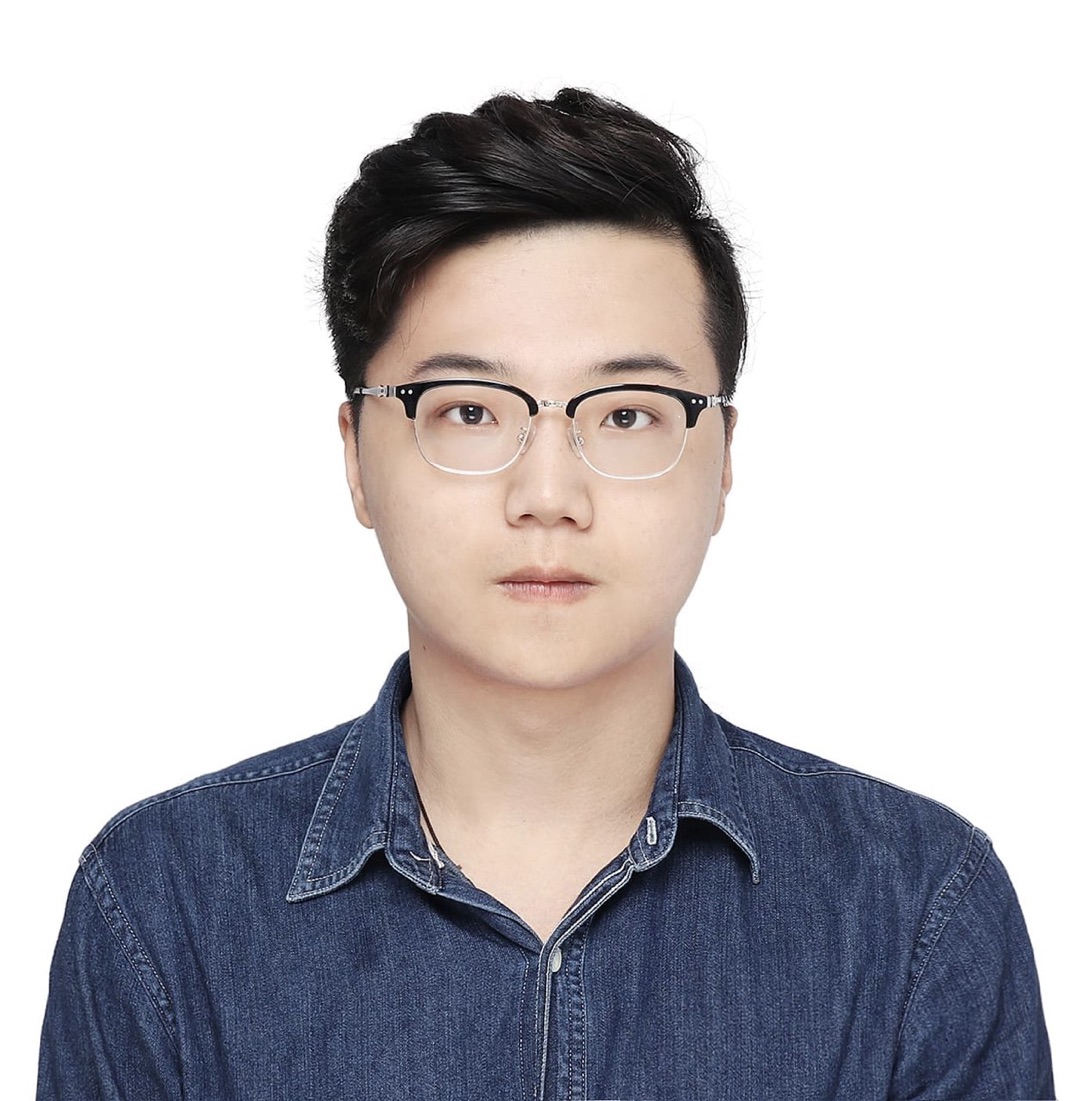}}]{Wenzhe Zhao}
		received a bachelor's degree and a master's degree from the South China University of Technology University, China. He is currently a Senior Researcher of Ads Multimedia Al at Tencent Data Platform. His research interests are in information retrieval, machine learning, deep learning, data mining, and computer vision. 
	\end{IEEEbiography}

	\begin{IEEEbiography}[{\includegraphics[width=1in,height=1.25in,clip,keepaspectratio]{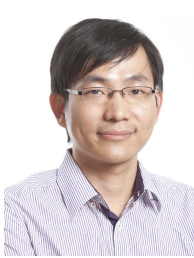}}]{Hongfa Wang} 
		received the BS degree from the Department of Mathematics, Southeast University, Nanjing, China, in 2005; and received the master’s degree in operation science and control theory from the Chinese Academy of Sciences, Beijing, China, in 2008. He is currently an Expert Researcher of Ads Multimedia AI at Tencent Data Platform. His research interests include  computer vision, machine learning, and pattern recognition.
	\end{IEEEbiography}

        \begin{IEEEbiography}[{\includegraphics[width=1in,height=1.25in,clip,keepaspectratio]{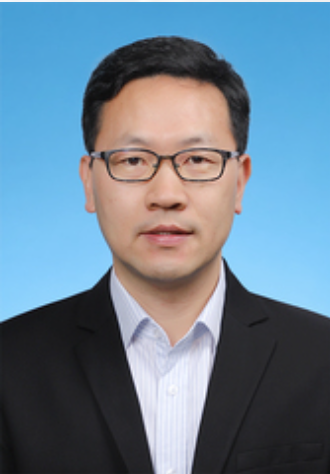}}]{Yujiu Yang} 
		received the Ph.D. degree from the Institute of Automation, the Chinese Academy of Sciences. He is currently an associate professor with Tsinghua Shenzhen International Graduate School, Tsinghua University. His research interests include machine learning, optimization theory, and their applications in natural language processing, visual content creation and understanding.
	\end{IEEEbiography}
 
        \begin{IEEEbiography}[{\includegraphics[width=1in,height=1.25in,clip,keepaspectratio]{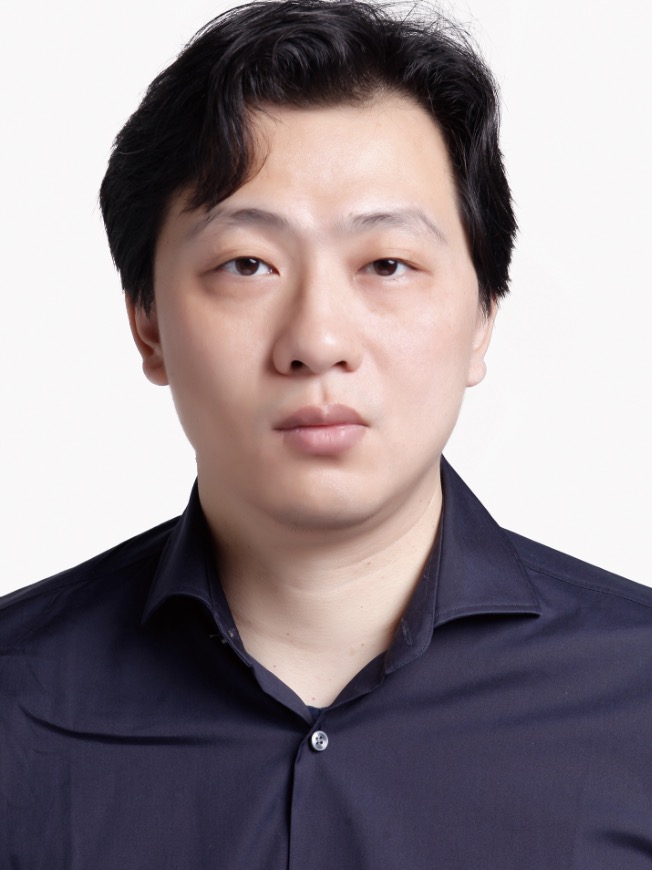}}]{Wei Liu} (M'14-SM'19-F'23) received the Ph.D. degree from Columbia University in 2012 in Electrical Engineering and Computer Science. He is currently a Distinguished Scientist of Tencent and the Director of Ads Multimedia AI at Tencent Data Platform. Prior to that, he has been a research staff member of IBM T. J. Watson Research Center, USA from 2012 to 2015. Dr. Liu has long been devoted to fundamental research and technological development in core fields of AI, including deep learning, machine learning, reinforcement learning, computer vision, information retrieval, big data, etc. To date, he has published extensively in these fields with more than 280 peer-reviewed technical papers, and also issued over 30 US patents. He currently serves on the editorial boards of internationally leading AI journals like IEEE Transactions on Pattern Analysis and Machine Intelligence (TPAMI), IEEE Transactions on Neural Networks and Learning Systems (TNNLS),  and IEEE Intelligent Systems. He is an Area Chair of top-tier computer science and AI conferences, e.g., NeurIPS, ICML, IEEE CVPR, IEEE ICCV, IJCAI, and AAAI. Dr. Liu is a Fellow of the IEEE, IAPR, and IMA, and an Elected Member of the ISI.    
	\end{IEEEbiography}
	
\end{document}